\documentclass[10pt,twocolumn,letterpaper]{article}

\usepackage{iccv}              
\usepackage{mathtools}

\newcommand{\teaser}{
\centering
\vspace{-1.2em}
{Project webpage:} \url{https://jianhongbai.github.io/ReCamMaster/}
\includegraphics[width=1.0\textwidth,trim=0em 0em 0em 0em,clip]{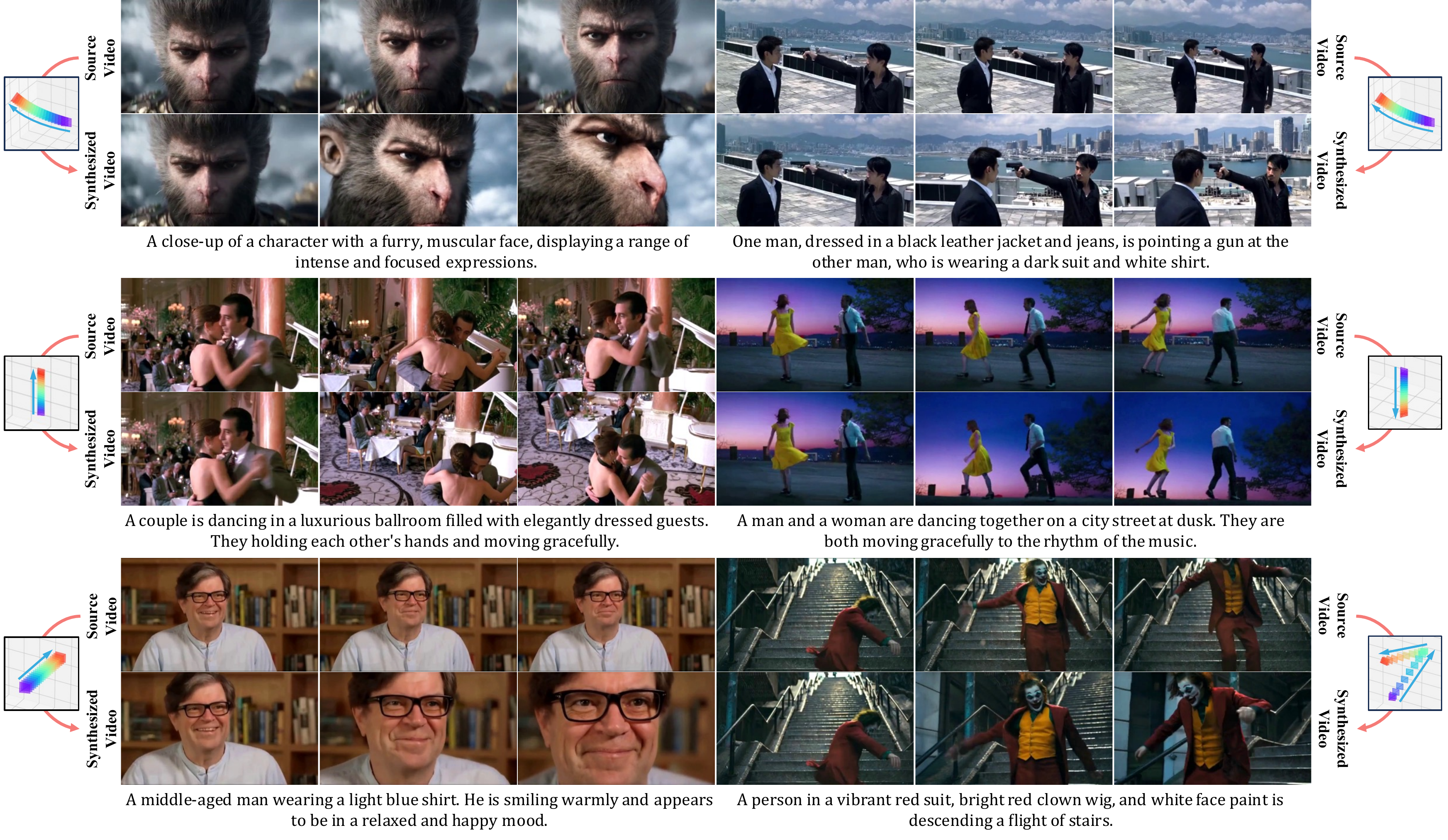}
\vspace{-2em}
\captionof{figure}{\textbf{Examples synthesized by ReCamMaster.} ReCamMaster re-shoots the source video with novel camera trajectories. We visualized the novel camera trajectories alongside the video frames. Video results are on our \href{https://jianhongbai.github.io/ReCamMaster/}{project page}. 
}
\vspace{0.7em}
\label{fig_1}
}

\definecolor{iccvblue}{rgb}{0.21,0.49,0.74}
\usepackage{url}
\usepackage{wrapfig}
\usepackage{graphicx}
\usepackage{enumitem}
\usepackage{mathtools}
\usepackage{array}
\usepackage{booktabs}
\usepackage{multirow}
\usepackage{makecell}
\usepackage{amssymb}
\usepackage{appendix}
\usepackage[accsupp]{axessibility}
\definecolor{citecolor}{HTML}{0071BC}
\definecolor{linkcolor}{HTML}{ED1C24}
\usepackage[pagebackref=false, breaklinks=true, colorlinks, citecolor=citecolor, linkcolor=linkcolor, bookmarks=false]{hyperref}
\graphicspath{{./figures/}}

\title{ReCamMaster: Camera-Controlled Generative Rendering from A Single Video}

\newcommand\nnfootnote[1]{
  \begin{NoHyper}
  \renewcommand\thefootnote{}\footnote{#1}
  \addtocounter{footnote}{-1}
  \end{NoHyper}
}

\begin{document}
\twocolumn[
\vspace{-0.5cm}
\author{\textbf{Jianhong Bai\textsuperscript{$\rm 1^{\ast}$}}, \textbf{Menghan Xia\textsuperscript{$\rm 2^{\dagger}$}}, \textbf{Xiao Fu\textsuperscript{\rm 3}}, \textbf{Xintao Wang\textsuperscript{\rm 2}}, \textbf{Lianrui Mu\textsuperscript{\rm 1}}, \textbf{Jinwen Cao\textsuperscript{\rm 2}},\\\textbf{Zuozhu Liu\textsuperscript{\rm 1}}, \textbf{Haoji Hu\textsuperscript{$\rm 1^{\dagger}$}}, \textbf{Xiang Bai\textsuperscript{\rm 4}}, \textbf{Pengfei Wan\textsuperscript{\rm 2}}, \textbf{Di Zhang\textsuperscript{\rm 2}} \\
\textsuperscript{\rm 1}Zhejiang University, \textsuperscript{\rm 2}Kling Team, Kuaishou Technology, \textsuperscript{\rm 3}CUHK, \textsuperscript{\rm 4}HUST\\
}
\maketitle
\teaser
]

\nnfootnote{$\ast$ Work done during an internship at Kling Team, Kuaishou Tech.} \nnfootnote{$\dagger$ Corresponding authors.}

\begin{abstract}
Camera control has been actively studied in text or image conditioned video generation tasks.
However, altering camera trajectories of a given video remains under-explored, despite its importance in the field of video creation. It is non-trivial due to the extra constraints of maintaining multiple-frame appearance and dynamic synchronization.
To address this, we present ReCamMaster, a camera-controlled generative video re-rendering framework that reproduces the dynamic scene of an input video at novel camera trajectories. The core innovation lies in harnessing the generative capabilities of pre-trained text-to-video models through a simple yet powerful video conditioning mechanism—its capability is often overlooked in current research.
To overcome the scarcity of qualified training data, we construct a comprehensive multi-camera synchronized video dataset using Unreal Engine 5, which is carefully curated to follow real-world filming characteristics, covering diverse scenes and camera movements. It helps the model generalize to in-the-wild videos.
Lastly, we further improve the robustness to diverse inputs through a meticulously designed training strategy.
Extensive experiments show that our method substantially outperforms existing state-of-the-art approaches. 
Our method also finds promising applications in video stabilization, super-resolution, and outpainting. Our code and dataset are publicly available at: \url{https://github.com/KwaiVGI/ReCamMaster}.
\end{abstract}

\clearpage
\section{Introduction}
\label{sec:intro}
Camera movement is a fundamental element in film production, profoundly shaping the audience's visual experience and conveying both emotional depth and narrative intentions.
For instance, a dolly-in shot can masterfully emphasize a character, building tension or guiding viewers' attention to crucial details. Similarly, crane shots excel at showcasing expansive landscapes, often employed at the opening to dramatically reveal the full scope of a setting.
Despite its artistic significance, achieving professional-level camera movement remains a challenge for amateur videographers, because of hardware limitations (such as stabilization issues in handheld recordings) and technical skill gaps.
To address this dilemma, we explore innovative approaches to modify camera trajectories in post-production, empowering creators with the ability to enhance their original footage by displaying the dynamic scenes with more compelling and polished camera trajectories when needed.

Previous research on camera-controlled generation has primarily focused on text-to-video (T2V) or image-to-video (I2V) generation \citep{motionctrl,cameractrl}. Recently, GCD \cite{gcd} pioneered camera-controlled video-to-video generation, achieving promising results on domain-specific videos synthesized by the Kubric simulator \cite{kubric}. However, its effectiveness on real-world videos is limited due to the narrow domain of training data and its inferior video conditioning mechanism.
As a concurrent work, ReCapture \cite{recapture} introduces a novel two-stage approach: first generating a camera-controlled anchor video using existing image-level multiview diffusion models, then optimizing the result by fine-tuning spatial and temporal LoRA \cite{lora} layers over the input and anchor videos respectively. While showing promising performance, its requirement for per-video optimization constrains practical applications.
Similarly, other approaches \cite{das, gs-dit} achieve video re-rendering through explicit 4D reconstruction followed by video post-optimization. However, their performance is significantly restricted by the challenge of single video-based 4D reconstruction techniques \cite{cotracker, spatialtracker}.

To address these challenges, we present ReCamMaster, a camera-controlled generative video re-rendering framework that can regenerate in-the-wild videos at novel camera trajectories.
The core innovation is to utilize the generative capabilities of pre-trained text-to-video models through an elegant yet powerful video conditioning mechanism. Compared to alternatives, our method shows superior performance in preserving both the visual and dynamic characteristics of the input video. Notably, this conditioning mechanism indeed exhibits remarkable potential as a versatile solution for conditional generation tasks, which is overlooked in current research.
Since no qualified training data are publicly available, we develop a large-scale multi-camera synchronized video dataset using Unreal Engine 5, which contains 136K realistic videos shot from 13.6K different dynamic scenes in 40 high-quality 3D environments with 122K different camera trajectories.
Particularly, it is carefully curated to simulate real-world filming characteristics, which is proven to bring an advantage to striking domain alignment to in-the-wild videos.
Lastly, we further improve the robustness of our model to diverse inputs through a meticulously designed training strategy.

Experimental results show that our method outperforms state-of-the-art approaches and strong baselines by a large margin. Ablation studies verified the effectiveness of our key designs. Furthermore, ReCamMaster demonstrates its potential in scenarios like video stabilization, video super-resolution, and video outpainting. Our contribution can be summarized as follows:
\begin{itemize}
    \item We introduce a high-quality multi-camera synchronized video dataset featuring diverse camera trajectories. This dataset has been publicly released to advance research in camera-controlled video generation, 4D reconstruction, and related fields.
    \item We conduct an in-depth investigation and validation of an effective video conditioning mechanism for text-to-video generation models. It significantly outperforms other alternatives employed in baseline methods.
    \item Extensive experiments show that our proposed ReCamMaster significantly advances the state-of-the-art in video recapturing. Furthermore, it finds promising applications across multiple real-world scenarios.
\end{itemize}
    
\section{Related Works}
\label{sec:related_works}

\paragraph{Camera-Controlled Video Generation.}
With the success of text-to-video generation models \citep{svd,  videocrafter2, cogvideox, snapvideo, hunyuanvideo}, the introduction of other conditional signals for controllable video generation has been widely studied \citep{dragnuwa, sparsectrl, makeyourvideo, fu20243dtrajmaster}. In camera-controlled video generation \cite{direct_a_video, vd3d, cami2v, zheng2025vidcraft3}, researchers aim to incorporate camera parameters into video generation models to control the viewpoint of the output video. AnimateDiff \citep{animatediff} introduces various motion LoRAs \citep{lora} to learn specific patterns of camera movements. MotionCtrl \citep{motionctrl} encodes 6DoF camera extrinsics and injects them into the diffusion model, fine-tuning on video-camera pair data to achieve video generation with arbitrary trajectory control. CameraCtrl \citep{cameractrl} further improves the accuracy and generalizability of single-sequence camera control with a dedicatedly designed camera encoder. CVD \citep{CVD} achieves multi-sequence camera control with the proposed cross-video synchronization module. AC3D \cite{ac3d} conducts an in-depth investigation into camera motion knowledge within diffusion transformers, achieving camera-controlled generation with enhanced visual quality. Meanwhile, training-free methods have also been explored \cite{hou2024training, hu2024motionmaster, ling2024motionclone}.

\paragraph{Video-to-Video Generation.}
Video-to-video generation \cite{wang2018video, wang2019few, mallya2020world} is explored in various tasks, including video editing \cite{rerender,videop2p}, video outpainting \cite{wang2024your,chen2024follow}, video super-resolution \cite{zhou2024upscale,xu2024videogigagan}, etc. Most relevant to our work is synthesizing novel viewpoint videos based on a given video and camera parameters \cite{gcd, recapture, gs-dit, das, ren2025gen3c}. GCD \cite{gcd} pioneered camera-controlled video-to-video generation by training a video generation model with paired video data created by the Kubric simulator \cite{kubric}. Although it performs well on in-domain data, GCD's generalization capability is limited by the significant domain gap between the training data and real-world videos. Recapture \cite{recapture} proposes a practical solution with per-video optimization. Several concurrent works \cite{yu2025trajectorycrafter, ren2025gen3c, das, trajattn, gs-dit}, achieve 4D-consistent video-to-video generation by extracting dynamic information from the original video using 3D point tracking \cite{cotracker,spatialtracker} and incorporating it as a condition into the video generator. This paradigm is promising when synchronized multi-view videos are unavailable. Nevertheless, the generation quality is limited by the accuracy of the point-tracking methods. In the field of 4D object generation, video-to-video generation can be achieved by 1) training a multi-view video generator \cite{sv4d, cat4d}, or 2) following a reconstruct-and-render pipeline via 4D reconstruction methods \cite{monst3r, megasam}. \cite{liu2024reconx, sun2024dimensionx} achieve 3D/4D scene reconstruction with videos generated with diffusion models.
In this paper, we focus on open-domain camera-controlled video generation. To achieve this, we first create a large multi-camera synchronized dataset with diverse camera trajectories. We also propose a novel video conditioning method to achieve better generation ability.

\section{Multi-Cam Video: A High-Quality Multi-Camera Synchronized Video Dataset}
\label{sec:data_collection}

To re-shoot input videos based on novel camera trajectories, paired video data is required for training the video generation model. Specifically, the training data should include multiple shots captured in the same scene simultaneously, allowing the model to learn 4D consistent generation. Acquiring such data in real-world scenarios is extremely costly, and publicly available multi-view synchronized datasets \cite{human36m, panoptic, shahroudy2016ntu, xu2024longvolcap, ego4d} are also limited by the diversity of scenes and constrained camera movements, making them unsuitable for our task. Therefore, we chose to use a rendering engine to generate the training data. The advantages of this approach are: 1) precise camera trajectories can be obtained, 2) complete synchronization in the time dimension can be achieved, and 3) the data volume can be more easily scaled up.
\begin{figure}[t]\centering
\includegraphics[width=0.47\textwidth]{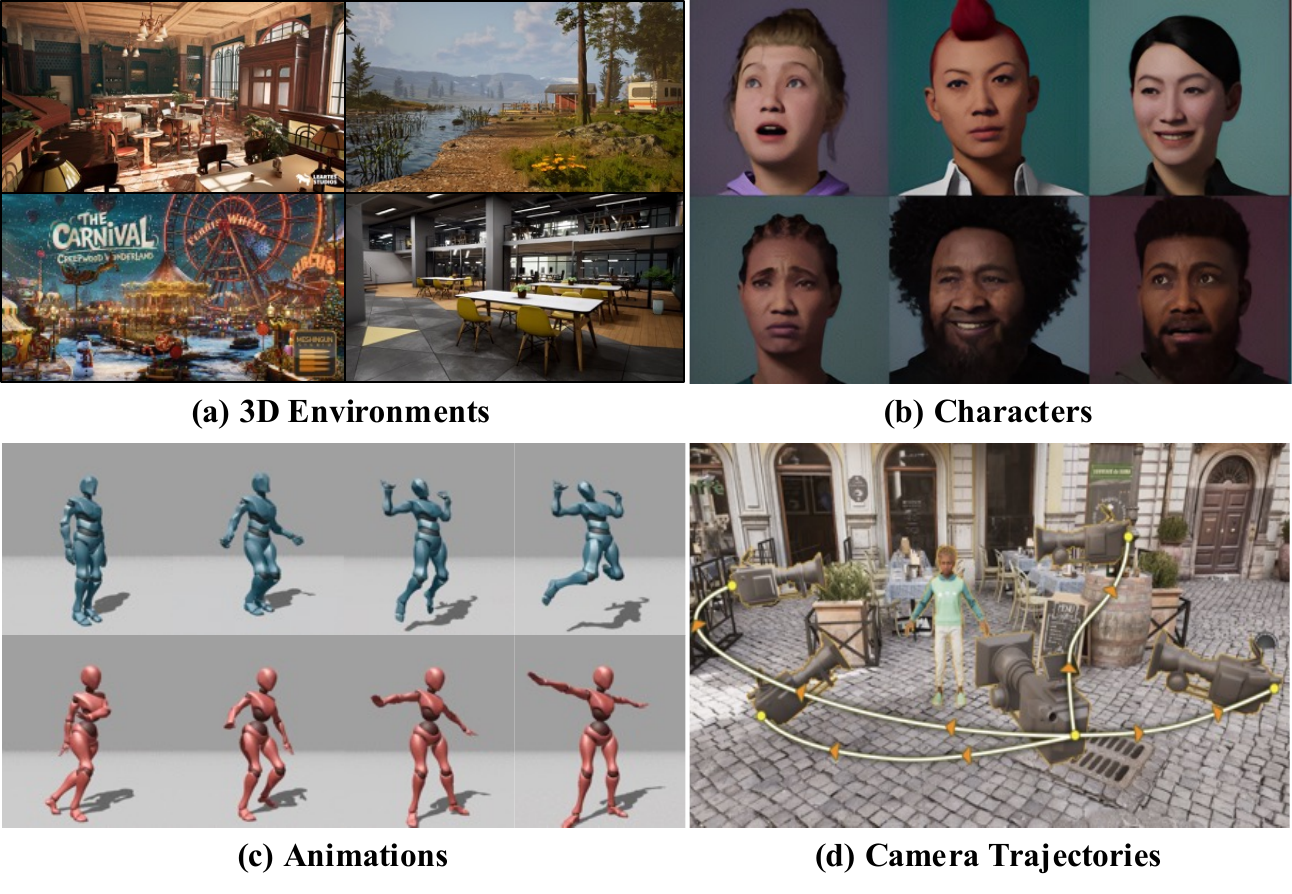}
\vspace{-0.3cm}
\caption{\textbf{Illustration of the dataset construction process.} We build the multi-camera synchronized training dataset by rendering in Unreal Engine 5. This is achieved using 3D environments, characters, animations collected from the internet, and our designed massive camera trajectories.}
    \vspace{-0.5cm}
    \label{fig_data}
\end{figure}

We build the entire data rendering pipeline in Unreal Engine 5 \cite{unrealengine5}. Specifically, we first collect multiple 3D environments as ``backgrounds". Then, we place animated characters within these environments as the ``main subjects" of the videos. We then position multiple cameras facing the subjects and moving along predefined trajectories to simulate the process of simultaneous shooting. This allows us to render datasets with synchronized cameras that include dynamic objects.
To scale up the data volume, we construct a set of camera movement rules for automatically batch generation of natural and diverse camera trajectories. Additionally, we randomly combined different characters and actions across different video sets.
In total, we obtain 136K visually-realistic videos shot from 13.6K different dynamic scenes in 40 high-quality 3D environments with 122K different camera trajectories.
Fig. \ref{fig_render_data_appendix} in the Appendix shows some of the rendered video frames.
In Appendix \ref{sec:appendix_data_ablation}, we conduct an ablation study comparing model performance when trained on a dataset with limited scenes and camera trajectories versus our comprehensive dataset. This highlights the significance of our high-quality dataset.
For more details about the dataset, please refer to Appendix \ref{sec:appendix_data_collection}.

\begin{figure*}[t]\centering
\includegraphics[width=0.95\textwidth]{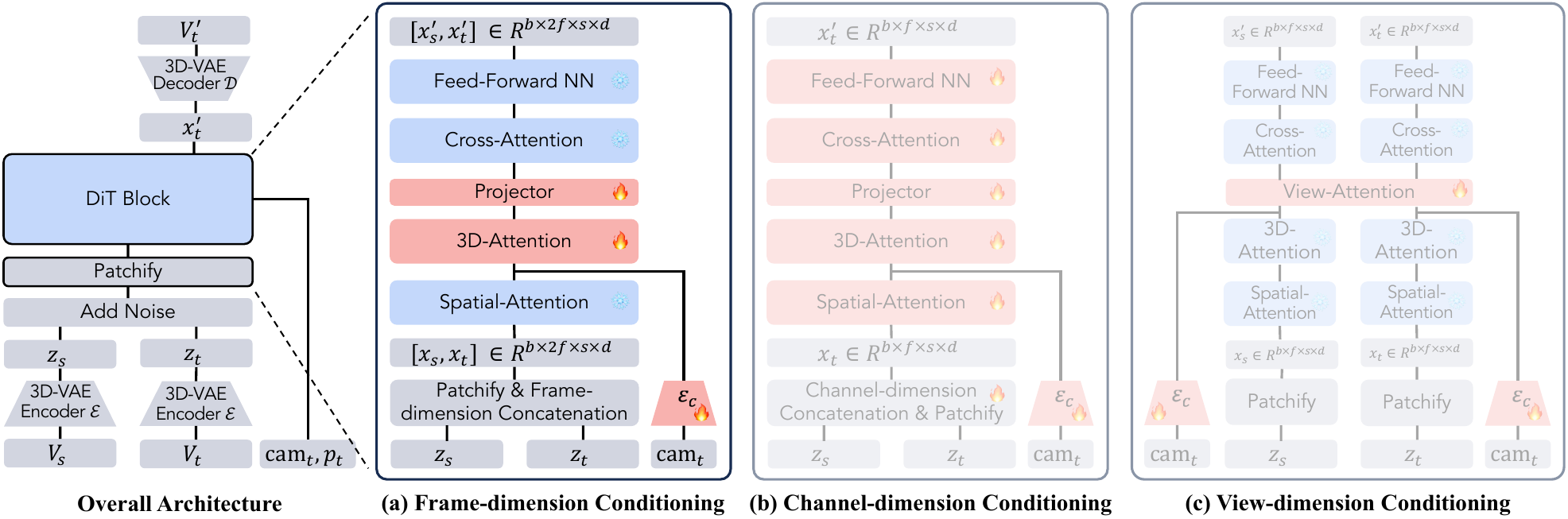}\vspace{-0.5em}
\caption{\textbf{Overview of ReCamMaster.} \textit{Left:} The training pipeline of ReCamMaster. A latent diffusion model is optimized to reconstruct the target video $V_t$, conditioned on the source video $V_s$, target camera pose $cam_t$, and target prompt $p_t$. \textit{Right:} Comparison of different video condition techniques. (a) Frame-dimension conditioning used in our paper; (b) Channel-dimension conditioning used in baseline methods \cite{gcd, gs-dit}; (c) View-dimension conditioning in \cite{syncammaster}. We omit the text prompt $p_t$ in (a)-(c) for simplicity.}
    \label{fig_pipe}
\vspace{-1.0em}
\end{figure*}

\section{Camera-Controlled Video Re-Generation}
\label{sec:method}

Given a source video $V_s \in \mathbb{R}^{f \times c \times h \times w}$, we aim to synthesize a target video $V_t \in \mathbb{R}^{f \times c \times h \times w}$ sharing the same dynamic scene but presenting at specified camera trajectories denoted by $\texttt{cam}_t \coloneqq [{R}, {t}] \in \mathbb{R}^{f\times3\times4}$. Particularly, $V_t$ should conform to the multiple-frame appearance and synchronized dynamics of $V_s$.
To achieve this, we propose to harness the generative capability of pre-trained text-to-video diffusion models~\citep{cogvideox,videocrafter2} by imposing dual conditions, \textit{i.e.}, the source video and target camera trajectories through a meticulously designed framework. The overview of the model is depicted in Fig.~\ref{fig_pipe}.

\subsection{Preliminary: Text-to-Video Base Model}
\label{sec:preliminary}

Our study is conducted over an internal pre-trained text-to-video foundation model. It is a latent video diffusion model, consisting of a 3D Variational Auto-Encoder (VAE)~\citep{kingma2014autoencoding} and a Transformer-based diffusion model (DiT)~\citep{dit}. Typically, each Transformer block is instantiated as a sequence of spatial attention, 3D (spatial-temporal) attention, and cross-attention modules. The generative model adopts Rectified Flow framework~\citep{scaling_esser_2024} for the noise schedule and denoising process. The forward process is defined as straight paths between data distribution and a standard normal distribution, i.e.
\begin{equation}\label{eq:forward}
    z_t = (1-t)z_0 + t\epsilon,
\end{equation}
where $\epsilon \in \mathcal{N}(0,I)$ and $t$ denotes the iterative timestep.
To solve the denoising processing, we define a mapping between samples $z_1$ from a noise distribution $p_1$ to samples
$z_0$ from a data distribution $p_0$ in terms of an ordinary differential equation (ODE), namely:
\begin{equation}\label{eq:ODE}
dz_t=v_{\Theta}(z_t,t)dt, 
\end{equation}
where the velocity $v$ is parameterized by the weights $\Theta$ of a neural network. For training, we regress a vector field $u_t$ that generates a probability path between $p_0$ and $p_1$ via Conditional Flow Matching~\citep{flow_lipman_2023}:
\begin{equation}\label{eq:loss}
    \mathcal{L}_{LCM}=\mathbb{E}_{t,p_t(z,\epsilon),p(\epsilon)} ||v_{\Theta}(z_t,t)-u_t(z_0|\epsilon)||_2^2,
\end{equation}
where $u_t(z,\epsilon):=\psi_t^{'}(\psi_t^{-1}(z|\epsilon)|\epsilon)$ with {$\psi(\cdot|\epsilon)$} denotes the function of Eq.~\ref{eq:forward}.
For inference, we employ Euler discretization for Eq. \ref{eq:ODE} and perform discretization over the timestep interval at $[0, 1]$, starting at $t=1$. We then processed with iterative sampling with:
\begin{equation}\label{eq:euler_sample}
z_t=z_{t-1} + v_{\Theta}(z_{t-1},t) * \Delta t.
\end{equation}

\subsection{Conditional Video Injection Mechanism}
\label{sec:method_video_condition}
To modify the camera trajectories of a given video using generative models, it is essential to condition the generation process on the source video. Our experiments reveal that the video injection mechanism is crucial for overall performance. We analyze and compare several widely-adopted video conditioning approaches from previous works \cite{gcd, gs-dit, syncammaster}, alongside our proposed simple yet powerful video injection mechanism. We provide an intuitive explanation of our design's superior performance and its potential impact.
\vspace{-0.5em}
\paragraph{Channel Dimension Conditioning.} Recent attempts \cite{gcd, gs-dit} incorporate the source video by concatenating the latent of the source video $x_s$ with the noised latent of the target video $x_t$ along the channel dimension, and add additional input channels to the input convolutional layer. To be specific, a variational autoencoder with encoder $\mathcal{E}$ is utilized to project the source video $V_s$ and target video $V_t$ to the latent space, $z_s=\mathcal{E}(V_s), z_t=\mathcal{E}(V_t)$, where $z_s, z_t \in \mathbb{R}^{b\times f \times c \times h \times w}$ are the latent of source video and target video with $f$ frames, $c$ channels, and spatial size of $h \times w$. Then, the source latent is concentrated with the noised target latent along the channel dimension, and patchified into $f \times h \times w$ tokens with several convolutional layers: 
\begin{equation}
    {x}_t = \texttt{patchify}([{z}_s, {z}_t]_{\text{channel-dim}}), 
\end{equation}
where $x_t \in \mathbb{R}^{b\times f \times s \times d}$, $s=h \times w$, $d$ is the channel dimension for the latent diffusion model. We denote this conditioning technique as the ``channel-dimension conditioning'' and illustrate in Fig. \ref{fig_pipe}(b).
\vspace{-0.5em}
\paragraph{View Dimension Conditioning.} To achieve multi-view video generation, \cite{syncammaster} proposes to introduce a plug-and-play module for feature aggregation across views. Given a source video latent $z_s=\mathcal{E}(V_s)$ and a target video latent $z_t=\mathcal{E}(V_t)$, and patchify and forward to the diffusion transformer respectively:
\begin{equation}
    {x}_s = \texttt{patchify}({z}_s), {x}_t = \texttt{patchify}({z}_t), 
\end{equation}
where $x_s, x_t \in \mathbb{R}^{b\times f \times s \times d}$. To achieve synchronization and content consistency across views, an additional attention layer (denoted as view-attention) is added in each basic transformer block, it performs self-attention between each frame in multiple views:
\begin{equation}
    \overline{{F}}_s^i, \overline{{F}}_t^i = \texttt{attn\_view}({F}_s^i, {F}_t^i), 
\end{equation}
where ${F}_s^i$ and ${F}_t^i$ are the $i$-th frame feature of $V_s$ and $V_t$ respectively, and $\overline{{F}}_s^i$ and $\overline{{F}}_t^i$ are the output features. We denote this conditioning technique as the ``view-dimension conditioning'' and illustrated in Fig. \ref{fig_pipe}(c).

\paragraph{Frame Dimension Conditioning (ours).} To achieve better synchronization and content consistency with the source video, we propose to concatenate the source video tokens with the target video tokens along the frame dimension:
\begin{equation}
\left\{
\begin{aligned}
    x_s &= \texttt{patchify}(z_s), \quad x_t = \texttt{patchify}(z_t), \\
    x_i &= [x_s, x_t]_{\text{frame-dim}},
\end{aligned}
\right.
\end{equation}
where $x_i \in \mathbb{R}^{b \times 2f \times s \times d}$ is the input of diffusion transformer. In other words, the input token number is doubled compared to the vanilla text-to-video generation process. Moreover, we do not introduce additional attention layers for feature aggregation across the source video and the target video, since self-attention is performed between all tokens in the 3D (spatial-temporal) attention layers. We denote this conditioning technique as the ``frame-dimension conditioning'' and illustrate in Fig. \ref{fig_pipe}(a).
\vspace{-1.0em}

\paragraph{Comparison and Discussion.}
Our experimental results demonstrate that the frame dimension conditioning approach provides substantial advantages in utilizing condition information effectively, as illustrated by the comparative analysis in Fig. \ref{fig_ablation_condition}.
We attribute this superior performance to the inherent flexibility of our frame dimension conditioning method, which enables spatio-temporal interaction between the conditional tokens and target tokens through all blocks of the base model. This approach provides a more robust mechanism for understanding the correlations between video pairs. While similar token concatenation strategies have proven effective in image-to-image tasks \cite{ominictrl}, their potential in video controllable generation has remained largely unexplored. Our study provides compelling evidence of the effectiveness, highlighting the potential of this underappreciated technique as a versatile solution for conditional generation tasks.

\subsection{Camera Pose Conditioning} 
To achieve camera-controlled video generation, a natural idea is to condition the model on the source and target video camera trajectories $\texttt{cam}_s$ and $\texttt{cam}_t$ to help the model better understand the 4D space. Unfortunately, during inference, even with state-of-the-art structure-from-motion (SfM) methods for camera parameter estimation, it is challenging to obtain accurate camera trajectory information for the input video \cite{ac3d}. Therefore, we alternate to only conditioning the model on the target camera $\texttt{cam}_t$, while relying on the model to interpret the camera trajectory of the input video.
Furthermore, we condition the model using camera extrinsics, specifically the rotation and translation matrices for each frame's camera. We chose not to include camera intrinsics as a condition because, while they are easily obtainable in the rendered training set, accurately estimating the intrinsics of real-world video cameras remains challenging. This limitation would complicate parameter provision in practical applications, as users are unlikely to have access to the source video's camera intrinsics. However, our method can be readily adapted to incorporate intrinsics as input with minimal modifications.

Given a $f$ frames target camera sequence $\texttt{cam}_t \in \mathbb{R}^{f \times (3 \times 4)}$, we project it to have the same channels with the video tokens through a learnable camera encoder $\mathcal{E}_c$ and add it to the visual features:
\begin{equation}
    F_i = F_o + \mathcal{E}_c(\texttt{cam}_t),
\end{equation}
where $F_o$ is the output feature of the spatial-attention layer, $F_i$ is the input feature of the 3D-attention layer, $\mathcal{E}_c$ is instantiated as a fully connected layer with an input dimension of 12 and an output dimension of $d$ in practice. The camera encoder is inserted into each transformer block for fine-grained camera control.

\begin{figure*}[t]\centering
    \includegraphics[width=0.98\textwidth]{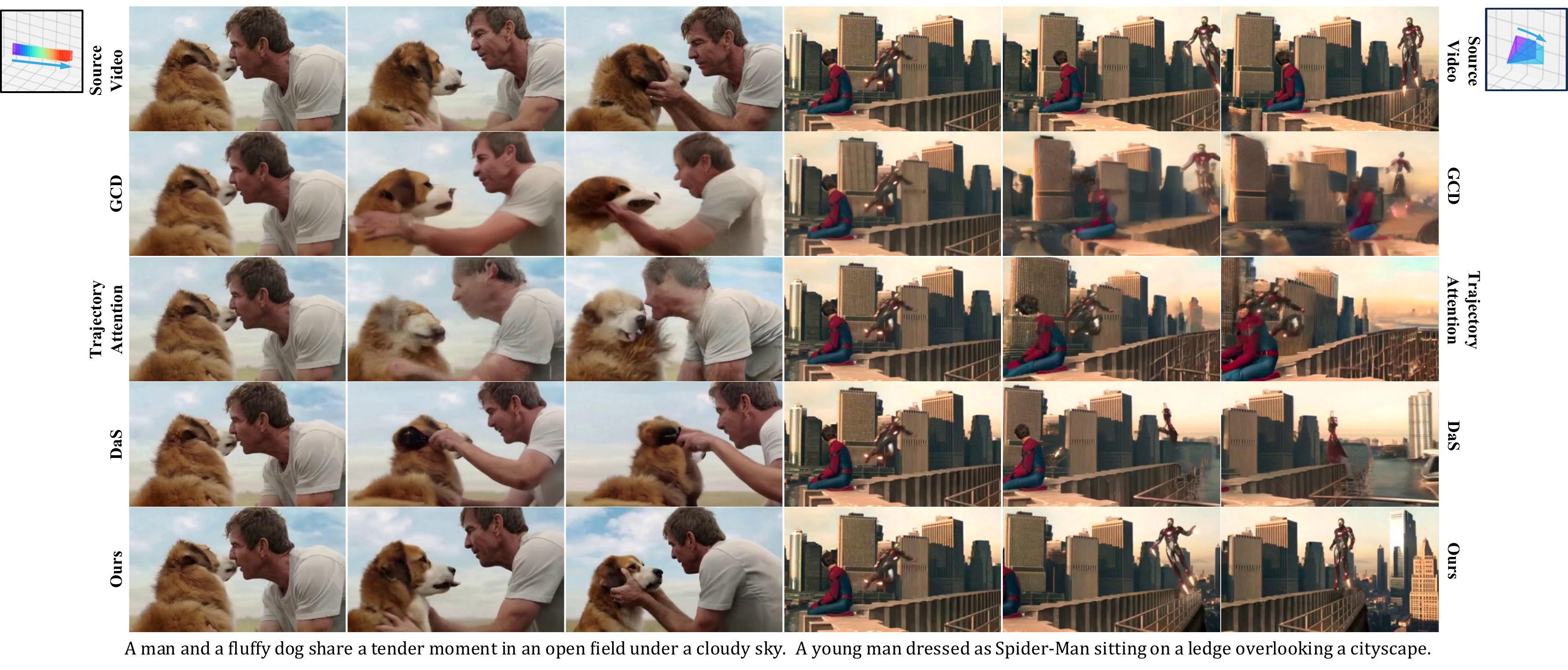}
    \vspace{-0.8em}
    \caption{\textbf{Comparison with state-of-the-art methods.} It shows that ReCamMaster generates videos that maintain appearance consistency and temporal synchronization with the source video.}
    \label{fig_exp_comparison}\vspace{-1.0em}
\end{figure*}

\subsection{Training Strategy}
\label{sec:training_strategy}
\paragraph{Enhancing Generalization Capabilities.}
Using the annotated dataset described in Sec. \ref{sec:data_collection}, ReCamMaster is trained to process input videos with target camera parameters. To preserve the base T2V model's native capability, we fine-tune only the camera encoder and 3D-attention layers while keeping other parameters frozen. To mitigate Unreal Engine's synthetic characteristics, we apply moderate noise (200-500 noise scheduling steps) to the conditional video latent during training, reducing the domain gap between synthetic and real-world data at inference.
\vspace{-0.5em}
\paragraph{Improving Generation Capability by Unifying Camera Control Tasks.}
To encourage content generation capability, we implement T2V camera-controlled generation \cite{motionctrl, cameractrl} with a 20\% probability and I2V camera-controlled generation \cite{camco} with a 20\% probability during training. Specifically, we replace the latent representations of all $f$ frames with Gaussian noise for T2V generation, and replace $f-1$ frames starting from the second frame for I2V generation. Our experiments demonstrate that this strategy prompts performance in synthesizing coherent objects invisible in the source video. As a byproduct, our model inherently supports T2V, I2V, and V2V camera-controlled generation, as shown in Fig. \ref{fig_unify_showcase}.

\section{Experimental Results}
\label{sec:results}

\begin{table*}[t]
	\begin{center}
            \vspace{-0.30cm}
		\caption{Quantitative comparison with state-of-the-art methods on visual quality, camera accuracy, and view synchronization.}
            \vspace{-0.30cm}
		\label{tab_quality_eval}
		\setlength\tabcolsep{3.2pt}
		\begin{tabular}{lcccc|cc|ccc}
			\toprule
			\multirow{2}{*}{Method} & \multicolumn{4}{c|}{Visual Quality} & \multicolumn{2}{c|}{Camera Accuracy} & \multicolumn{3}{c}{View Synchronization}\\ 
                \cmidrule(r){2-10}
                & \makecell[c]{FID $\downarrow$} & \makecell[c]{FVD $\downarrow$} & \makecell[c]{CLIP-T $\uparrow$} & \makecell[c]{CLIP-F $\uparrow$} & RotErr $\downarrow$& TransErr $\downarrow$ & {Mat. Pix.(K) $\uparrow$} & {FVD-V $\downarrow$} & {CLIP-V $\uparrow$}\\
                \midrule 
                GCD & 72.83 & 367.32 & 32.86 & 95.66 & 2.27 & 5.51 & 639.39 & 365.75 & 85.92 \\
                Trajectory-Attention & 69.21 & 276.06 & 33.43 & 96.52 & 2.18 & 5.32 & 619.13 & 256.30 & 88.65 \\
                DaS & 63.25 & 159.60 & 33.05 & 98.32 & 1.45 & 5.59 & 633.53 & 154.25 & 87.33 \\

                \midrule
                ReCamMaster & \textbf{57.10} & \textbf{122.74} & \textbf{34.53} & \textbf{98.74} & \textbf{1.22} & \textbf{4.85} & \textbf{906.03} & \textbf{90.38} & \textbf{90.36} \\
			\bottomrule
		\end{tabular}
	\end{center}
        \vspace{-0.5cm}
\end{table*}
\begin{table*}[t]
	\begin{center}
		\caption{Quantitative comparison with state-of-the-art methods on VBench \cite{vbench} metrics.}
            \vspace{-0.30cm}
		\label{tab_quality_eval_vbench}
		\setlength\tabcolsep{8.3pt}
		\begin{tabular}{lccccccc}
			\toprule
                {Method} & \makecell[c]{Aesthetic\\Quality $\uparrow$} & \makecell[c]{Imaging\\Quality $\uparrow$} & \makecell[c]{Temporal\\Flickering $\uparrow$} & \makecell[c]{Motion\\Smoothness $\uparrow$} & \makecell[c]{Subject\\Consistency $\uparrow$} & \makecell[c]{Background\\Consistency $\uparrow$}\\
                \midrule 
                GCD& 38.21 & 41.56 & 95.81 & 98.37 & 88.94 & 92.00 \\
                Trajectory-Attention& 38.50 & 51.00 & 95.52 & 98.21 & 90.60 & 92.83 \\
                DaS& 39.86 & 51.55 & \textbf{97.44} & 99.14 & 90.34 & 92.03 \\
                \midrule
                ReCamMaster& \textbf{42.70} & \textbf{53.97} & 97.36 & \textbf{99.28} & \textbf{92.05} & \textbf{93.83} \\

			\bottomrule
		\end{tabular}
	\end{center}
        \vspace{-0.5cm}
\end{table*}
\begin{table*}[t]
	\begin{center}
            \vspace{-0.2cm}
		\caption{Quantitative comparison on the video conditioning strategy.}
            \vspace{-0.30cm}
		\label{tab_ablation_conditioning}
		\setlength\tabcolsep{3.7pt}
		\begin{tabular}{lcccc|cc|ccc}
			\toprule
			\multirow{2}{*}{Method} & \multicolumn{4}{c|}{Visual Quality} & \multicolumn{2}{c|}{Camera Accuracy} & \multicolumn{3}{c}{View Synchronization}\\ 
                \cmidrule(r){2-10}
                & \makecell[c]{FID $\downarrow$} & \makecell[c]{FVD $\downarrow$} & \makecell[c]{CLIP-T $\uparrow$} & \makecell[c]{CLIP-F $\uparrow$} & RotErr $\downarrow$& TransErr $\downarrow$ & {Mat. Pix.(K) $\uparrow$} & {FVD-V $\downarrow$} & {CLIP-V $\uparrow$}\\
                \midrule 
                Channel dim. & 74.09 & 187.94 & 33.70 & 98.67 & 1.28 & 4.98 & 521.10 & 148.51 & 84.62 \\
                View dim. & 80.51 & 194.47 & \textbf{34.66} & 98.71 & 1.42 & 5.77 & 573.92 & 177.68 & 83.40 \\
                Frame dim. (ours) & \textbf{57.10} & \textbf{122.74} & 34.53 & \textbf{98.74} & \textbf{1.22} & \textbf{4.85} & \textbf{906.03} & \textbf{90.38} & \textbf{90.36} \\
			\bottomrule
		\end{tabular}
	\end{center}
        \vspace{-0.8cm}
\end{table*}
\subsection{Experiment Settings}
\vspace{0.35cm}
\noindent\textbf{Implementation Details.}
We train ReCamMaster on the rendered dataset introduced in Sec. \ref{sec:data_collection}. During training, we randomly select 2 cameras from each 10 synchronized cameras, designating one as the source video and the other as the target video. We use \{source video $V_s$, target camera $\texttt{cam}_t$, target prompt $p_t$\} as conditions input to the model, to reconstruct the target video $V_t$ as introduced in Eq. \ref{eq:loss}. We train the model for 10K steps at the resolution of 384x672 with a learning rate of 0.0001, batch size 40. The camera encoder and the projector are zero-initialized.

\vspace{0.4cm}
\noindent\textbf{Evaluation Metrics.}
We mainly evaluate the proposed method in terms of camera accuracy, source-target synchronization, and visual quality. For camera accuracy, we use
GLOMAP \cite{pan2024global} to extract the camera pose sequence of the generated videos, and calculate the rotation error and translation error, denoted as RotErr and TransErr respectively \citep{cameractrl}. In terms of synchronization, we utilize the state-of-the-art image matching method GIM \citep{gim} to calculate the number of matching pixels with confidence greater than the threshold, denoted as Mat. Pix.. Furthermore, we calculate the FVD-V score in SV4D \citep{sv4d}, and the average CLIP similarity between source and target frames at the same timestamp, denoted as CLIP-V \citep{CVD}. For visual quality, we divide it into fidelity, coherence with text, and temporal consistency, and quantify them with Fréchet Image Distance \citep{fid} (FID) and Fréchet Video Distance \citep{fvd} (FVD), CLIP-T, and CLIP-F, respectively. CLIP-T refers to the average CLIP \citep{clip} similarity of each frame and its corresponding text prompt, and CLIP-F is the average CLIP similarity of adjacent frames. We also evaluate our method on the widely used VBench \cite{vbench} metrics.

\vspace{0.4cm}
\noindent\textbf{Evaluation Set.} We construct the evaluation set with 1000 random videos from WebVid \cite{webvid} and 10 different camera trajectories, including pan, tilt, vertical translation, zoom in, zoom out, and horizontal arc trajectories. Note that since we automatically generated 122K different camera trajectories when constructing the training set, ReCamMaster can support a variety of camera trajectories as input. However, since the baseline method was trained on specific basic trajectories, we opted to use trajectories that closely align with the baseline for comparison.
\subsection{Comparison with State-of-the-Art Methods}

\paragraph{Baselines.} We compare the proposed ReCamMaster with state-of-the-art camera-controlled video-to-video generation methods \cite{gcd, trajattn, das}. GCD \cite{gcd} pioneered camera-controlled video-to-video generation by training a video generation model with paired data created by the Kubric simulator \cite{kubric}. Although it performs well on in-domain data, GCD's generalization capability is limited by its inferior video conditioning method and the significant domain gap between the training data and real-world videos. Trajectory-Attention \cite{trajattn} and DaS \cite{das} extract dynamic information from the source video using 3D point tracking and incorporate it as a condition into the video generator.
\paragraph{Qualitative Results.}
We present synthesized examples of ReCamMaster in Fig. \ref{fig_1} (additional examples in Fig. \ref{fig_appendix_ours1} of Appendix \ref{sec:appendix_results}). Please visit our \href{https://jianhongbai.github.io/ReCamMaster/}{project page} for more videos. ReCamMaster demonstrates the ability to: 1) generate consistent content from novel camera trajectories of the same scene; 2) achieve excellent temporal synchronization with the source video; and 3) generate plausible video content in areas not visible in the original video. Note that, due to the use of 122K different random trajectories during training, ReCamMaster also supports complex trajectories as input, such as zigzag paths. Please refer to the results on the project page. In the paper, to visually capture the camera movement trajectory more clearly, we do not showcase results generated along complex trajectories.

We compare ReCamMaster with state-of-the-art methods in Fig. \ref{fig_exp_comparison} and Fig. \ref{fig_appendix_compare1}. It's observed that baselines produce content with notable artifacts and temporal desynchronization. Similar phenomena were noted when training ReCamMaster with other video conditioning techniques (as shown in Fig. \ref{fig_ablation_condition}), highlighting the importance of the novel video conditioning method proposed in the paper.

\paragraph{Quantitative Results.}
We quantitatively evaluate ReCamMaster against baselines using various automatic metrics, the summarized results are in Tab. \ref{tab_quality_eval} and \ref{tab_quality_eval_vbench}. For visual quality assessment, we used the widely adopted FID, FVD, and CLIP metrics as shown in Tab. \ref{tab_quality_eval}, along with the VBench metrics in Tab. \ref{tab_quality_eval_vbench}. In terms of camera trajectory accuracy, we calculate the rotation error and translation error by following previous work in camera-controlled T2V and I2V generation \cite{cameractrl, camco}. For view synchronization, we calculate the clip similarity score and FVD between video frames of different viewpoints within one scene, denoted as CLIP-V and FVD-V. It's observed that ReCamMaster outperforms baselines across multiple dimensions of metrics.

\subsection{More Analysis and Ablation Studies}
\paragraph{Ablation on Video Conditioning Techniques.}
\begin{figure}[t]\centering
    \includegraphics[width=0.47\textwidth]{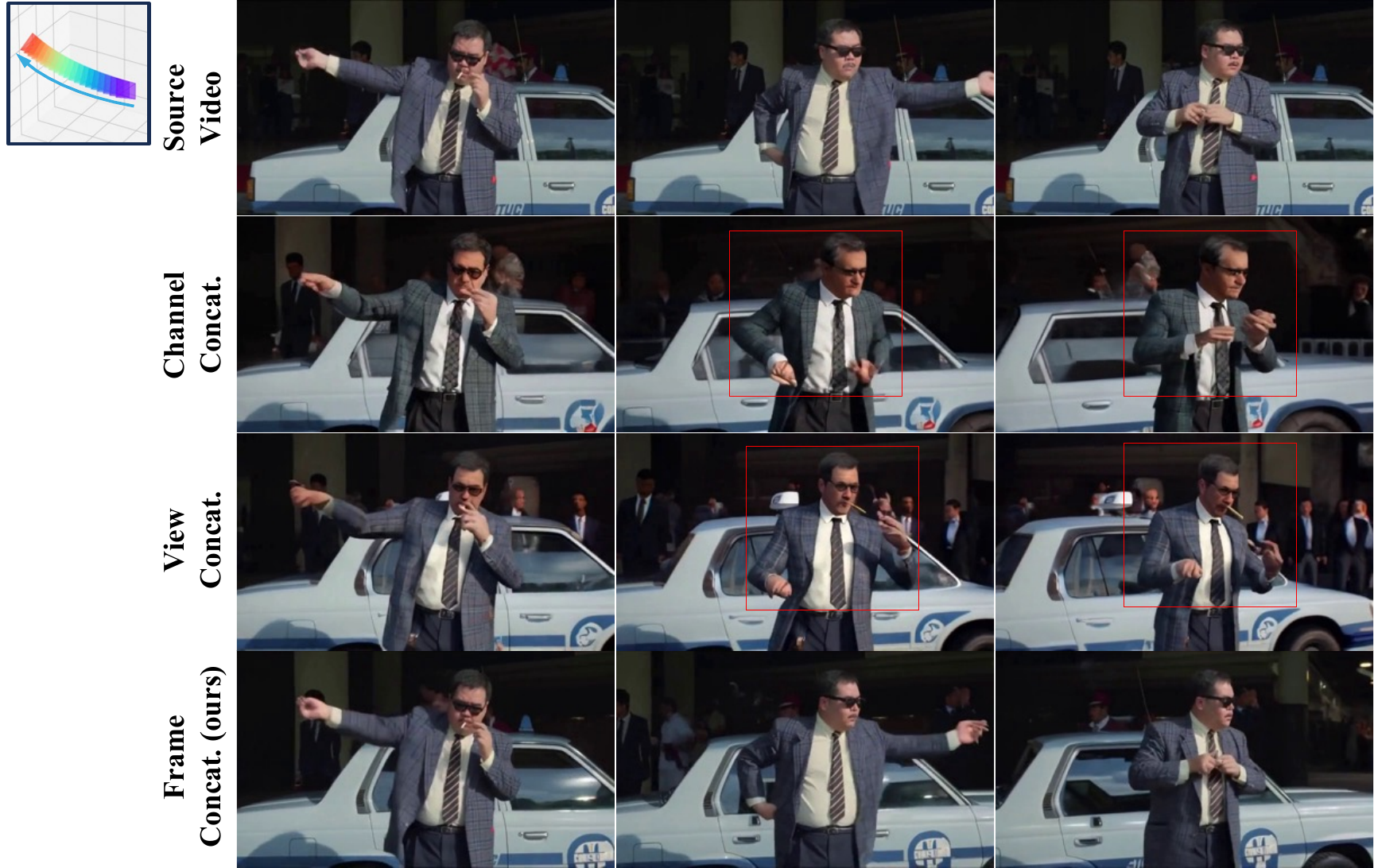}\vspace{-0.5em}
    \caption{\textbf{Ablation on video conditioning techniques.} We compared the channel-/view- concatenation schemes proposed by previous methods and frame-concatenation in ReCamMaster. We observed that both channel-conditioning and view-conditioning schemes suffer from significant artifacts, content inconsistency, and asynchronous dynamics with respect to the original video.}
    \label{fig_ablation_condition}\vspace{-1.0em}
\end{figure}
In Sec. \ref{sec:method_video_condition}, we introduce a novel video conditioning scheme that concatenates the tokens of a source video with the target video tokens along the frame dimension. To verify its effectiveness, we compare our ``frame concatenation" technique with the ``channel concatenation" used in baseline methods \cite{gcd,gs-dit} and the ``view concatenation" in \cite{syncammaster}. We performed both qualitative and quantitative comparisons, as shown in Fig. \ref{fig_ablation_condition} and Tab. \ref{tab_ablation_conditioning}. We only altered the video conditioning method while keeping the camera injection method, training data, and training steps unchanged. The results clearly show that our conditioning technique significantly enhances the model's performance. In the second and third rows of Fig. \ref{fig_ablation_condition}, the generated videos exhibit the rendering style of the training dataset, and the hand movements become unsynchronized with the original video. In contrast, our method preserves the person's identity and maintains synchronization even during rapid and complex movements.
\vspace{-0.5cm}

\paragraph{The Effectiveness of the Training Strategies.}
In Section \ref{sec:training_strategy}, we enhance ReCamMaster's robustness with improved training strategies. We fine-tune only the spatial-temporal (3D) attention layers and freeze other parameters. Additionally, we drop $f$ frames and $f-1$ frames of the source video with a 20\% probability to improve generation capability. Table \ref{tab_ablation_training} shows the improvement using visual quality metrics. The `baseline' is vanilla training with reconstruction loss. Results indicate that all techniques improve visual quality, with their combination yielding the best performance. Dropping the source latent during training also allows our method to support camera-controlled T2V, I2V, and V2V generation simultaneously, as shown in Fig. \ref{fig_unify_showcase}.

\vspace{-0.2cm}
\section{Applications of ReCamMaster}

\begin{table}[t]
	\begin{center}
        \small
		\caption{Ablation on our training strategies.}
            \vspace{-0.30cm}
		\label{tab_ablation_training}
		\setlength\tabcolsep{5pt}
		\begin{tabular}{lcccc}
			\toprule
                {Method} & \makecell[c]{FID $\downarrow$} & \makecell[c]{FVD $\downarrow$} & \makecell[c]{Aesthetic\\Quality $\uparrow$} & \makecell[c]{Imaging\\Quality $\uparrow$} \\
                \midrule 
                Baseline& 66.67 & 171.80 & 40.02 & 51.93 \\
                \midrule
                + Add noise &  65.17 & 164.04 & 40.36 & 52.22\\
                + 3D-Attn. tuning& 59.47 & 132.58 & \textbf{43.08} & 52.80 \\
                + Drop latent & 62.39 & 149.52 & 41.47 & 52.65 \\
                + All & \textbf{57.10} & \textbf{122.74} & 42.70 & \textbf{53.97} \\

			\bottomrule
		\end{tabular}
        \vspace{-0.3cm}
	\end{center}
        \vspace{-0.6cm}
\end{table}

\begin{figure}[t]
    \centering
    \includegraphics[width=0.46\textwidth]{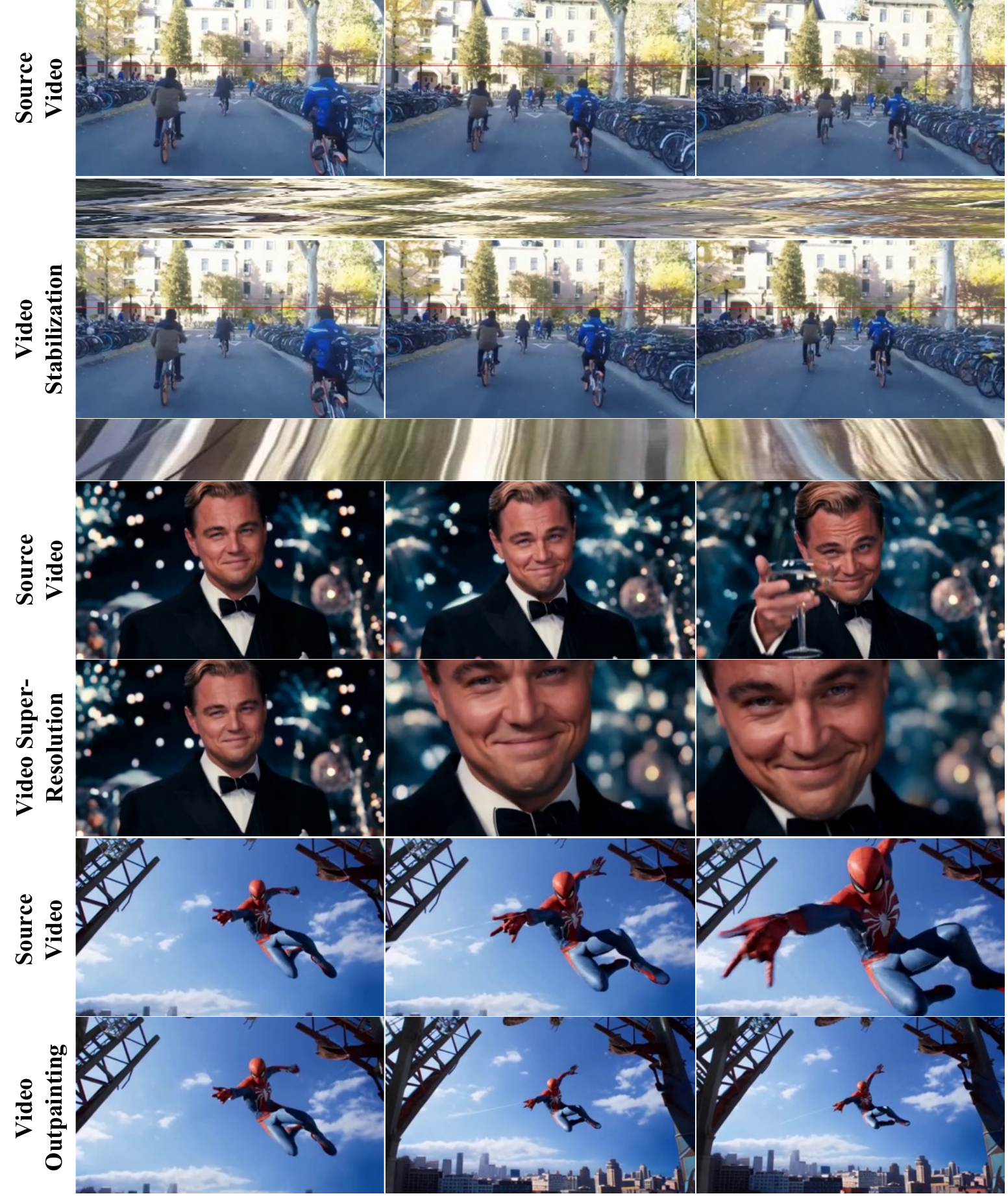}
    \vspace{-0.5em}
    \caption{\textbf{Applications of ReCamMaster.} \textit{From top to bottom:} video stabilization, video super-resolution, and video outpainting.}
    \label{fig_application}
    \vspace{-1.0em}
\end{figure}

We find that ReCamMaster also has promising performances in several traditional tasks, as shown in Fig. \ref{fig_application}.

\noindent\textbf{Video Stabilization.}
It is usually not easy to obtain stable videos when recording video while moving. Video stabilization techniques \cite{wang2023video} aim to smooth out camera movements to produce easy-to-watch videos, which can be achieved by adjusting the camera trajectories via ReCamMaster. We evaluate our method using unsteady videos from the DeepStab \cite{deepstab} dataset. It can be observed that the model stabilizes the video while preserving the content from the original video. Best viewed on the \href{https://jianhongbai.github.io/ReCamMaster/}{project page}.

\noindent\textbf{Video Super-Resolution.\footnote{It's not entirely equivalent to the super-resolution task, as ReCamMaster only enhances the resolution of patches in the central region.}}
Thanks to the generation capability of the diffusion model, we can input `zoom-in' camera trajectories into ReCamMaster to achieve video local super-resolution, as more details are observed in Fig. \ref{fig_application}.

\noindent\textbf{Video Outpainting.}
Similarly, we can input zoom-out trajectories to achieve video outpainting. As shown in the last row of Fig. \ref{fig_application} that areas not visible in the original video, such as feet and the ground, have been generated.

\section{Conclusion and Limitations}
\label{sec:conclusion}
In this paper, we propose ReCamMaster to reproduce dynamic scenes from input videos with new camera trajectories. We develop an innovative video conditioning technique to enhance pre-trained text-to-video models and curate a large-scale multi-camera synchronized video dataset using Unreal Engine 5, covering diverse scenes and camera movements. Our method also shows promise in video stabilization, super-resolution, and outpainting. There are nevertheless some limitations. Concatenating source and target video tokens improves generation quality but increases computational demands. Additionally, ReCamMaster inherits limitations from the pre-trained T2V models, such as less effective hand generation, as shown in Fig. \ref{fig_failure}.
\clearpage

\section*{Acknowledgments}
We thank Jinwen Cao, Yisong Guo, Haowen Ji, Jichao Wang, and Yi Wang from Kuaishou Technology for their invaluable help in constructing the multi-view video dataset.
Jianhong Bai would like to thank Quande Liu and Xiaoyu Shi for fruitful discussions. We thank Qinghe Wang and Yawen Luo for their help in method evaluation.
{
    \small
    \bibliographystyle{ieeenat_fullname}
    \bibliography{main}

\begin{thebibliography}{69}
\providecommand{\natexlab}[1]{#1}
\providecommand{\url}[1]{\texttt{#1}}
\expandafter\ifx\csname urlstyle\endcsname\relax
  \providecommand{\doi}[1]{doi: #1}\else
  \providecommand{\doi}{doi: \begingroup \urlstyle{rm}\Url}\fi

\bibitem[Bahmani et~al.(2024{\natexlab{a}})Bahmani, Skorokhodov, Qian, Siarohin, Menapace, Tagliasacchi, Lindell, and Tulyakov]{ac3d}
Sherwin Bahmani, Ivan Skorokhodov, Guocheng Qian, Aliaksandr Siarohin, Willi Menapace, Andrea Tagliasacchi, David~B Lindell, and Sergey Tulyakov.
\newblock Ac3d: Analyzing and improving 3d camera control in video diffusion transformers.
\newblock \emph{arXiv preprint arXiv:2411.18673}, 2024{\natexlab{a}}.

\bibitem[Bahmani et~al.(2024{\natexlab{b}})Bahmani, Skorokhodov, Siarohin, Menapace, Qian, Vasilkovsky, Lee, Wang, Zou, Tagliasacchi, et~al.]{vd3d}
Sherwin Bahmani, Ivan Skorokhodov, Aliaksandr Siarohin, Willi Menapace, Guocheng Qian, Michael Vasilkovsky, Hsin-Ying Lee, Chaoyang Wang, Jiaxu Zou, Andrea Tagliasacchi, et~al.
\newblock Vd3d: Taming large video diffusion transformers for 3d camera control.
\newblock \emph{arXiv preprint arXiv:2407.12781}, 2024{\natexlab{b}}.

\bibitem[Bai et~al.(2024)Bai, Xia, Wang, Yuan, Fu, Liu, Hu, Wan, and Zhang]{syncammaster}
Jianhong Bai, Menghan Xia, Xintao Wang, Ziyang Yuan, Xiao Fu, Zuozhu Liu, Haoji Hu, Pengfei Wan, and Di Zhang.
\newblock Syncammaster: Synchronizing multi-camera video generation from diverse viewpoints.
\newblock \emph{arXiv preprint arXiv:2412.07760}, 2024.

\bibitem[Bain et~al.(2021)Bain, Nagrani, Varol, and Zisserman]{webvid}
Max Bain, Arsha Nagrani, G{\"u}l Varol, and Andrew Zisserman.
\newblock Frozen in time: A joint video and image encoder for end-to-end retrieval.
\newblock In \emph{IEEE International Conference on Computer Vision}, 2021.

\bibitem[Bian et~al.(2025)Bian, Huang, Shi, Li, Wang, and Li]{gs-dit}
Weikang Bian, Zhaoyang Huang, Xiaoyu Shi, Yijin Li, Fu-Yun Wang, and Hongsheng Li.
\newblock Gs-dit: Advancing video generation with pseudo 4d gaussian fields through efficient dense 3d point tracking.
\newblock \emph{arXiv preprint arXiv:2501.02690}, 2025.

\bibitem[Blattmann et~al.(2023)Blattmann, Dockhorn, Kulal, Mendelevitch, Kilian, Lorenz, Levi, English, Voleti, Letts, et~al.]{svd}
Andreas Blattmann, Tim Dockhorn, Sumith Kulal, Daniel Mendelevitch, Maciej Kilian, Dominik Lorenz, Yam Levi, Zion English, Vikram Voleti, Adam Letts, et~al.
\newblock Stable video diffusion: Scaling latent video diffusion models to large datasets.
\newblock \emph{arXiv preprint arXiv:2311.15127}, 2023.

\bibitem[Chen et~al.(2024{\natexlab{a}})Chen, Zhang, Cun, Xia, Wang, Weng, and Shan]{videocrafter2}
Haoxin Chen, Yong Zhang, Xiaodong Cun, Menghan Xia, Xintao Wang, Chao Weng, and Ying Shan.
\newblock Videocrafter2: Overcoming data limitations for high-quality video diffusion models.
\newblock In \emph{Proceedings of the IEEE/CVF Conference on Computer Vision and Pattern Recognition}, pages 7310--7320, 2024{\natexlab{a}}.

\bibitem[Chen et~al.(2024{\natexlab{b}})Chen, Ma, Wang, Yuan, Zhao, Tian, Wang, Min, Chen, and Liu]{chen2024follow}
Qihua Chen, Yue Ma, Hongfa Wang, Junkun Yuan, Wenzhe Zhao, Qi Tian, Hongmei Wang, Shaobo Min, Qifeng Chen, and Wei Liu.
\newblock Follow-your-canvas: Higher-resolution video outpainting with extensive content generation.
\newblock \emph{arXiv preprint arXiv:2409.01055}, 2024{\natexlab{b}}.

\bibitem[Esser et~al.(2024)Esser, Kulal, Blattmann, Entezari, M{\"{u}}ller, Saini, Levi, Lorenz, Sauer, Boesel, Podell, Dockhorn, English, and Rombach]{scaling_esser_2024}
Patrick Esser, Sumith Kulal, Andreas Blattmann, Rahim Entezari, Jonas M{\"{u}}ller, Harry Saini, Yam Levi, Dominik Lorenz, Axel Sauer, Frederic Boesel, Dustin Podell, Tim Dockhorn, Zion English, and Robin Rombach.
\newblock Scaling rectified flow transformers for high-resolution image synthesis.
\newblock In \emph{International Conference on Machine Learning (ICML)}, 2024.

\bibitem[et. al.(2024)]{hunyuanvideo}
Weijie~Kong et. al.
\newblock Hunyuanvideo: A systematic framework for large video generative models, 2024.

\bibitem[Fu et~al.(2024)Fu, Liu, Wang, Peng, Xia, Shi, Yuan, Wan, Zhang, and Lin]{fu20243dtrajmaster}
Xiao Fu, Xian Liu, Xintao Wang, Sida Peng, Menghan Xia, Xiaoyu Shi, Ziyang Yuan, Pengfei Wan, Di Zhang, and Dahua Lin.
\newblock 3dtrajmaster: Mastering 3d trajectory for multi-entity motion in video generation.
\newblock \emph{arXiv preprint arXiv:2412.07759}, 2024.

\bibitem[Games(2022)]{unrealengine5}
Epic Games.
\newblock Unreal engine 5.
\newblock \url{https://www.unrealengine.com/en-US/unreal-engine-5}, 2022.

\bibitem[Grauman et~al.(2024)Grauman, Westbury, Torresani, Kitani, Malik, Afouras, Ashutosh, Baiyya, Bansal, Boote, et~al.]{ego4d}
Kristen Grauman, Andrew Westbury, Lorenzo Torresani, Kris Kitani, Jitendra Malik, Triantafyllos Afouras, Kumar Ashutosh, Vijay Baiyya, Siddhant Bansal, Bikram Boote, et~al.
\newblock Ego-exo4d: Understanding skilled human activity from first-and third-person perspectives.
\newblock In \emph{Proceedings of the IEEE/CVF Conference on Computer Vision and Pattern Recognition}, pages 19383--19400, 2024.

\bibitem[Greff et~al.(2022)Greff, Belletti, Beyer, Doersch, Du, Duckworth, Fleet, Gnanapragasam, Golemo, Herrmann, Kipf, Kundu, Lagun, Laradji, Liu, Meyer, Miao, Nowrouzezahrai, Oztireli, Pot, Radwan, Rebain, Sabour, Sajjadi, Sela, Sitzmann, Stone, Sun, Vora, Wang, Wu, Yi, Zhong, and Tagliasacchi]{kubric}
Klaus Greff, Francois Belletti, Lucas Beyer, Carl Doersch, Yilun Du, Daniel Duckworth, David~J Fleet, Dan Gnanapragasam, Florian Golemo, Charles Herrmann, Thomas Kipf, Abhijit Kundu, Dmitry Lagun, Issam Laradji, Hsueh-Ti~(Derek) Liu, Henning Meyer, Yishu Miao, Derek Nowrouzezahrai, Cengiz Oztireli, Etienne Pot, Noha Radwan, Daniel Rebain, Sara Sabour, Mehdi S.~M. Sajjadi, Matan Sela, Vincent Sitzmann, Austin Stone, Deqing Sun, Suhani Vora, Ziyu Wang, Tianhao Wu, Kwang~Moo Yi, Fangcheng Zhong, and Andrea Tagliasacchi.
\newblock Kubric: a scalable dataset generator.
\newblock 2022.

\bibitem[Gu et~al.(2025)Gu, Yan, Lu, Li, Dou, Si, Dong, Liu, Lin, Liu, et~al.]{das}
Zekai Gu, Rui Yan, Jiahao Lu, Peng Li, Zhiyang Dou, Chenyang Si, Zhen Dong, Qifeng Liu, Cheng Lin, Ziwei Liu, et~al.
\newblock Diffusion as shader: 3d-aware video diffusion for versatile video generation control.
\newblock \emph{arXiv preprint arXiv:2501.03847}, 2025.

\bibitem[Guo et~al.(2023{\natexlab{a}})Guo, Yang, Rao, Agrawala, Lin, and Dai]{sparsectrl}
Yuwei Guo, Ceyuan Yang, Anyi Rao, Maneesh Agrawala, Dahua Lin, and Bo Dai.
\newblock Sparsectrl: Adding sparse controls to text-to-video diffusion models.
\newblock \emph{arXiv preprint arXiv:2311.16933}, 2023{\natexlab{a}}.

\bibitem[Guo et~al.(2023{\natexlab{b}})Guo, Yang, Rao, Liang, Wang, Qiao, Agrawala, Lin, and Dai]{animatediff}
Yuwei Guo, Ceyuan Yang, Anyi Rao, Zhengyang Liang, Yaohui Wang, Yu Qiao, Maneesh Agrawala, Dahua Lin, and Bo Dai.
\newblock Animatediff: Animate your personalized text-to-image diffusion models without specific tuning.
\newblock \emph{arXiv preprint arXiv:2307.04725}, 2023{\natexlab{b}}.

\bibitem[He et~al.(2024)He, Xu, Guo, Wetzstein, Dai, Li, and Yang]{cameractrl}
Hao He, Yinghao Xu, Yuwei Guo, Gordon Wetzstein, Bo Dai, Hongsheng Li, and Ceyuan Yang.
\newblock Cameractrl: Enabling camera control for text-to-video generation.
\newblock \emph{arXiv preprint arXiv:2404.02101}, 2024.

\bibitem[Heusel et~al.(2017)Heusel, Ramsauer, Unterthiner, Nessler, and Hochreiter]{fid}
Martin Heusel, Hubert Ramsauer, Thomas Unterthiner, Bernhard Nessler, and Sepp Hochreiter.
\newblock Gans trained by a two time-scale update rule converge to a local nash equilibrium.
\newblock \emph{Advances in neural information processing systems}, 30, 2017.

\bibitem[Hou et~al.(2024)Hou, Wei, Zeng, and Chen]{hou2024training}
Chen Hou, Guoqiang Wei, Yan Zeng, and Zhibo Chen.
\newblock Training-free camera control for video generation.
\newblock \emph{arXiv preprint arXiv:2406.10126}, 2024.

\bibitem[Hu et~al.(2021)Hu, Shen, Wallis, Allen-Zhu, Li, Wang, Wang, and Chen]{lora}
Edward~J Hu, Yelong Shen, Phillip Wallis, Zeyuan Allen-Zhu, Yuanzhi Li, Shean Wang, Lu Wang, and Weizhu Chen.
\newblock Lora: Low-rank adaptation of large language models.
\newblock \emph{arXiv preprint arXiv:2106.09685}, 2021.

\bibitem[Hu et~al.(2024)Hu, Zhang, Yi, Wang, Huang, Weng, Wang, and Ma]{hu2024motionmaster}
Teng Hu, Jiangning Zhang, Ran Yi, Yating Wang, Hongrui Huang, Jieyu Weng, Yabiao Wang, and Lizhuang Ma.
\newblock Motionmaster: Training-free camera motion transfer for video generation.
\newblock \emph{arXiv preprint arXiv:2404.15789}, 2024.

\bibitem[Huang et~al.(2024)Huang, He, Yu, Zhang, Si, Jiang, Zhang, Wu, Jin, Chanpaisit, et~al.]{vbench}
Ziqi Huang, Yinan He, Jiashuo Yu, Fan Zhang, Chenyang Si, Yuming Jiang, Yuanhan Zhang, Tianxing Wu, Qingyang Jin, Nattapol Chanpaisit, et~al.
\newblock Vbench: Comprehensive benchmark suite for video generative models.
\newblock In \emph{Proceedings of the IEEE/CVF Conference on Computer Vision and Pattern Recognition}, pages 21807--21818, 2024.

\bibitem[Ionescu et~al.(2013)Ionescu, Papava, Olaru, and Sminchisescu]{human36m}
Catalin Ionescu, Dragos Papava, Vlad Olaru, and Cristian Sminchisescu.
\newblock Human3. 6m: Large scale datasets and predictive methods for 3d human sensing in natural environments.
\newblock \emph{IEEE transactions on pattern analysis and machine intelligence}, 36\penalty0 (7):\penalty0 1325--1339, 2013.

\bibitem[Joo et~al.(2015)Joo, Liu, Tan, Gui, Nabbe, Matthews, Kanade, Nobuhara, and Sheikh]{panoptic}
Hanbyul Joo, Hao Liu, Lei Tan, Lin Gui, Bart Nabbe, Iain Matthews, Takeo Kanade, Shohei Nobuhara, and Yaser Sheikh.
\newblock Panoptic studio: A massively multiview system for social motion capture.
\newblock In \emph{Proceedings of the IEEE international conference on computer vision}, pages 3334--3342, 2015.

\bibitem[Karaev et~al.(2024)Karaev, Rocco, Graham, Neverova, Vedaldi, and Rupprecht]{cotracker}
Nikita Karaev, Ignacio Rocco, Benjamin Graham, Natalia Neverova, Andrea Vedaldi, and Christian Rupprecht.
\newblock Cotracker: It is better to track together.
\newblock In \emph{European Conference on Computer Vision}, pages 18--35. Springer, 2024.

\bibitem[Kingma and Welling(2014)]{kingma2014autoencoding}
Diederik~P Kingma and Max Welling.
\newblock Auto-encoding variational bayes.
\newblock In \emph{International Conference on Learning Representations (ICLR)}, 2014.

\bibitem[Kuang et~al.(2024)Kuang, Cai, He, Xu, Li, Guibas, and Wetzstein]{CVD}
Zhengfei Kuang, Shengqu Cai, Hao He, Yinghao Xu, Hongsheng Li, Leonidas Guibas, and Gordon Wetzstein.
\newblock Collaborative video diffusion: Consistent multi-video generation with camera control.
\newblock \emph{arXiv preprint arXiv:2405.17414}, 2024.

\bibitem[Li et~al.(2024)Li, Tucker, Cole, Wang, Jin, Ye, Kanazawa, Holynski, and Snavely]{megasam}
Zhengqi Li, Richard Tucker, Forrester Cole, Qianqian Wang, Linyi Jin, Vickie Ye, Angjoo Kanazawa, Aleksander Holynski, and Noah Snavely.
\newblock Megasam: Accurate, fast, and robust structure and motion from casual dynamic videos.
\newblock \emph{arXiv preprint arXiv:2412.04463}, 2024.

\bibitem[Ling et~al.(2024)Ling, Bu, Zhang, Dong, Zang, Wu, Chen, Wang, and Jin]{ling2024motionclone}
Pengyang Ling, Jiazi Bu, Pan Zhang, Xiaoyi Dong, Yuhang Zang, Tong Wu, Huaian Chen, Jiaqi Wang, and Yi Jin.
\newblock Motionclone: Training-free motion cloning for controllable video generation.
\newblock \emph{arXiv preprint arXiv:2406.05338}, 2024.

\bibitem[Lipman et~al.(2023)Lipman, Chen, Ben{-}Hamu, Nickel, and Le]{flow_lipman_2023}
Yaron Lipman, Ricky T.~Q. Chen, Heli Ben{-}Hamu, Maximilian Nickel, and Matthew Le.
\newblock Flow matching for generative modeling.
\newblock In \emph{International Conference on Learning Representations (ICLR)}, 2023.

\bibitem[Liu et~al.(2024{\natexlab{a}})Liu, Sun, Wang, Wang, Sun, Ye, Zhang, and Duan]{liu2024reconx}
Fangfu Liu, Wenqiang Sun, Hanyang Wang, Yikai Wang, Haowen Sun, Junliang Ye, Jun Zhang, and Yueqi Duan.
\newblock Reconx: Reconstruct any scene from sparse views with video diffusion model.
\newblock \emph{arXiv preprint arXiv:2408.16767}, 2024{\natexlab{a}}.

\bibitem[Liu et~al.(2024{\natexlab{b}})Liu, Zhang, Li, Lin, and Jia]{videop2p}
Shaoteng Liu, Yuechen Zhang, Wenbo Li, Zhe Lin, and Jiaya Jia.
\newblock Video-p2p: Video editing with cross-attention control.
\newblock In \emph{Proceedings of the IEEE/CVF Conference on Computer Vision and Pattern Recognition}, pages 8599--8608, 2024{\natexlab{b}}.

\bibitem[Mallya et~al.(2020)Mallya, Wang, Sapra, and Liu]{mallya2020world}
Arun Mallya, Ting-Chun Wang, Karan Sapra, and Ming-Yu Liu.
\newblock World-consistent video-to-video synthesis.
\newblock In \emph{Computer Vision--ECCV 2020: 16th European Conference, Glasgow, UK, August 23--28, 2020, Proceedings, Part VIII 16}, pages 359--378. Springer, 2020.

\bibitem[Menapace et~al.(2024)Menapace, Siarohin, Skorokhodov, Deyneka, Chen, Kag, Fang, Stoliar, Ricci, Ren, et~al.]{snapvideo}
Willi Menapace, Aliaksandr Siarohin, Ivan Skorokhodov, Ekaterina Deyneka, Tsai-Shien Chen, Anil Kag, Yuwei Fang, Aleksei Stoliar, Elisa Ricci, Jian Ren, et~al.
\newblock Snap video: Scaled spatiotemporal transformers for text-to-video synthesis.
\newblock In \emph{Proceedings of the IEEE/CVF Conference on Computer Vision and Pattern Recognition}, pages 7038--7048, 2024.

\bibitem[Pan et~al.(2024)Pan, Bar{\'a}th, Pollefeys, and Sch{\"o}nberger]{pan2024global}
Linfei Pan, D{\'a}niel Bar{\'a}th, Marc Pollefeys, and Johannes~L Sch{\"o}nberger.
\newblock Global structure-from-motion revisited.
\newblock In \emph{European Conference on Computer Vision}, pages 58--77. Springer, 2024.

\bibitem[Peebles and Xie(2023)]{dit}
William Peebles and Saining Xie.
\newblock Scalable diffusion models with transformers.
\newblock In \emph{Proceedings of the IEEE/CVF International Conference on Computer Vision}, pages 4195--4205, 2023.

\bibitem[Radford et~al.(2021)Radford, Kim, Hallacy, Ramesh, Goh, Agarwal, Sastry, Askell, Mishkin, Clark, et~al.]{clip}
Alec Radford, Jong~Wook Kim, Chris Hallacy, Aditya Ramesh, Gabriel Goh, Sandhini Agarwal, Girish Sastry, Amanda Askell, Pamela Mishkin, Jack Clark, et~al.
\newblock Learning transferable visual models from natural language supervision.
\newblock In \emph{International conference on machine learning}, pages 8748--8763. PMLR, 2021.

\bibitem[Ren et~al.(2025)Ren, Shen, Huang, Ling, Lu, Nimier-David, Müller, Keller, Fidler, and Gao]{ren2025gen3c}
Xuanchi Ren, Tianchang Shen, Jiahui Huang, Huan Ling, Yifan Lu, Merlin Nimier-David, Thomas Müller, Alexander Keller, Sanja Fidler, and Jun Gao.
\newblock Gen3c: 3d-informed world-consistent video generation with precise camera control.
\newblock In \emph{Proceedings of the IEEE/CVF Conference on Computer Vision and Pattern Recognition}, 2025.

\bibitem[Shahroudy et~al.(2016)Shahroudy, Liu, Ng, and Wang]{shahroudy2016ntu}
Amir Shahroudy, Jun Liu, Tian-Tsong Ng, and Gang Wang.
\newblock Ntu rgb+ d: A large scale dataset for 3d human activity analysis.
\newblock In \emph{Proceedings of the IEEE conference on computer vision and pattern recognition}, pages 1010--1019, 2016.

\bibitem[Shen et~al.(2024)Shen, Cai, Yin, Müller, Li, Wang, Chen, and Wang]{gim}
Xuelun Shen, Zhipeng Cai, Wei Yin, Matthias Müller, Zijun Li, Kaixuan Wang, Xiaozhi Chen, and Cheng Wang.
\newblock Gim: Learning generalizable image matcher from internet videos.
\newblock In \emph{The Twelfth International Conference on Learning Representations}, 2024.

\bibitem[Sun et~al.(2024)Sun, Chen, Liu, Chen, Duan, Zhang, and Wang]{sun2024dimensionx}
Wenqiang Sun, Shuo Chen, Fangfu Liu, Zilong Chen, Yueqi Duan, Jun Zhang, and Yikai Wang.
\newblock Dimensionx: Create any 3d and 4d scenes from a single image with controllable video diffusion.
\newblock \emph{arXiv preprint arXiv:2411.04928}, 2024.

\bibitem[Tan et~al.(2024)Tan, Liu, Yang, Xue, and Wang]{ominictrl}
Zhenxiong Tan, Songhua Liu, Xingyi Yang, Qiaochu Xue, and Xinchao Wang.
\newblock Ominicontrol: Minimal and universal control for diffusion transformer.
\newblock \emph{arXiv preprint arXiv:2411.15098}, 2024.

\bibitem[Unterthiner et~al.(2019)Unterthiner, van Steenkiste, Kurach, Marinier, Michalski, and Gelly]{fvd}
Thomas Unterthiner, Sjoerd van Steenkiste, Karol Kurach, Rapha{\"e}l Marinier, Marcin Michalski, and Sylvain Gelly.
\newblock Fvd: A new metric for video generation.
\newblock \emph{Openreview}, 2019.

\bibitem[Van~Hoorick et~al.(2024)Van~Hoorick, Wu, Ozguroglu, Sargent, Liu, Tokmakov, Dave, Zheng, and Vondrick]{gcd}
Basile Van~Hoorick, Rundi Wu, Ege Ozguroglu, Kyle Sargent, Ruoshi Liu, Pavel Tokmakov, Achal Dave, Changxi Zheng, and Carl Vondrick.
\newblock Generative camera dolly: Extreme monocular dynamic novel view synthesis.
\newblock In \emph{European Conference on Computer Vision}, pages 313--331. Springer, 2024.

\bibitem[Wang et~al.(2024{\natexlab{a}})Wang, Wu, Huang, Shi, Shen, Song, Liu, and Li]{wang2024your}
Fu-Yun Wang, Xiaoshi Wu, Zhaoyang Huang, Xiaoyu Shi, Dazhong Shen, Guanglu Song, Yu Liu, and Hongsheng Li.
\newblock Be-your-outpainter: Mastering video outpainting through input-specific adaptation.
\newblock In \emph{European Conference on Computer Vision}, pages 153--168. Springer, 2024{\natexlab{a}}.

\bibitem[Wang et~al.(2018{\natexlab{a}})Wang, Yang, Lin, Zhang, Shamir, Lu, and Hu]{deepstab}
Miao Wang, Guo-Ye Yang, Jin-Kun Lin, Song-Hai Zhang, Ariel Shamir, Shao-Ping Lu, and Shi-Min Hu.
\newblock Deep online video stabilization with multi-grid warping transformation learning.
\newblock \emph{IEEE Transactions on Image Processing}, 28\penalty0 (5):\penalty0 2283--2292, 2018{\natexlab{a}}.

\bibitem[Wang et~al.(2018{\natexlab{b}})Wang, Liu, Zhu, Liu, Tao, Kautz, and Catanzaro]{wang2018video}
Ting-Chun Wang, Ming-Yu Liu, Jun-Yan Zhu, Guilin Liu, Andrew Tao, Jan Kautz, and Bryan Catanzaro.
\newblock Video-to-video synthesis.
\newblock \emph{arXiv preprint arXiv:1808.06601}, 2018{\natexlab{b}}.

\bibitem[Wang et~al.(2019)Wang, Liu, Tao, Liu, Kautz, and Catanzaro]{wang2019few}
Ting-Chun Wang, Ming-Yu Liu, Andrew Tao, Guilin Liu, Jan Kautz, and Bryan Catanzaro.
\newblock Few-shot video-to-video synthesis.
\newblock \emph{arXiv preprint arXiv:1910.12713}, 2019.

\bibitem[Wang et~al.(2023)Wang, Huang, Jiang, Liu, Shang, and Miao]{wang2023video}
Yiming Wang, Qian Huang, Chuanxu Jiang, Jiwen Liu, Mingzhou Shang, and Zhuang Miao.
\newblock Video stabilization: A comprehensive survey.
\newblock \emph{Neurocomputing}, 516:\penalty0 205--230, 2023.

\bibitem[Wang et~al.(2024{\natexlab{b}})Wang, Yuan, Wang, Li, Chen, Xia, Luo, and Shan]{motionctrl}
Zhouxia Wang, Ziyang Yuan, Xintao Wang, Yaowei Li, Tianshui Chen, Menghan Xia, Ping Luo, and Ying Shan.
\newblock Motionctrl: A unified and flexible motion controller for video generation.
\newblock In \emph{ACM SIGGRAPH 2024 Conference Papers}, pages 1--11, 2024{\natexlab{b}}.

\bibitem[Wu et~al.(2024)Wu, Gao, Poole, Trevithick, Zheng, Barron, and Holynski]{cat4d}
Rundi Wu, Ruiqi Gao, Ben Poole, Alex Trevithick, Changxi Zheng, Jonathan~T Barron, and Aleksander Holynski.
\newblock Cat4d: Create anything in 4d with multi-view video diffusion models.
\newblock \emph{arXiv preprint arXiv:2411.18613}, 2024.

\bibitem[Xiao et~al.(2024)Xiao, Wang, Zhang, Xue, Peng, Shen, and Zhou]{spatialtracker}
Yuxi Xiao, Qianqian Wang, Shangzhan Zhang, Nan Xue, Sida Peng, Yujun Shen, and Xiaowei Zhou.
\newblock Spatialtracker: Tracking any 2d pixels in 3d space.
\newblock In \emph{Proceedings of the IEEE/CVF Conference on Computer Vision and Pattern Recognition}, pages 20406--20417, 2024.

\bibitem[Xiao et~al.(2025)Xiao, Ouyang, Zhou, Yang, Yang, Si, and Pan]{trajattn}
Zeqi Xiao, Wenqi Ouyang, Yifan Zhou, Shuai Yang, Lei Yang, Jianlou Si, and Xingang Pan.
\newblock Trajectory attention for fine-grained video motion control.
\newblock In \emph{The Thirteenth International Conference on Learning Representations}, 2025.

\bibitem[Xie et~al.(2024)Xie, Yao, Voleti, Jiang, and Jampani]{sv4d}
Yiming Xie, Chun-Han Yao, Vikram Voleti, Huaizu Jiang, and Varun Jampani.
\newblock Sv4d: Dynamic 3d content generation with multi-frame and multi-view consistency.
\newblock \emph{arXiv preprint arXiv:2407.17470}, 2024.

\bibitem[Xing et~al.(2024)Xing, Xia, Liu, Zhang, Zhang, He, Liu, Chen, Cun, Wang, et~al.]{makeyourvideo}
Jinbo Xing, Menghan Xia, Yuxin Liu, Yuechen Zhang, Yong Zhang, Yingqing He, Hanyuan Liu, Haoxin Chen, Xiaodong Cun, Xintao Wang, et~al.
\newblock Make-your-video: Customized video generation using textual and structural guidance.
\newblock \emph{IEEE Transactions on Visualization and Computer Graphics}, 2024.

\bibitem[Xu et~al.(2024{\natexlab{a}})Xu, Nie, Liu, Liu, Kautz, Wang, and Vahdat]{camco}
Dejia Xu, Weili Nie, Chao Liu, Sifei Liu, Jan Kautz, Zhangyang Wang, and Arash Vahdat.
\newblock Camco: Camera-controllable 3d-consistent image-to-video generation.
\newblock \emph{arXiv preprint arXiv:2406.02509}, 2024{\natexlab{a}}.

\bibitem[Xu et~al.(2024{\natexlab{b}})Xu, Park, Zhang, Zhou, Shechtman, Liu, Huang, and Liu]{xu2024videogigagan}
Yiran Xu, Taesung Park, Richard Zhang, Yang Zhou, Eli Shechtman, Feng Liu, Jia-Bin Huang, and Difan Liu.
\newblock Videogigagan: Towards detail-rich video super-resolution.
\newblock \emph{arXiv preprint arXiv:2404.12388}, 2024{\natexlab{b}}.

\bibitem[Xu et~al.(2024{\natexlab{c}})Xu, Xu, Yu, Peng, Sun, Bao, and Zhou]{xu2024longvolcap}
Zhen Xu, Yinghao Xu, Zhiyuan Yu, Sida Peng, Jiaming Sun, Hujun Bao, and Xiaowei Zhou.
\newblock Representing long volumetric video with temporal gaussian hierarchy.
\newblock \emph{ACM Transactions on Graphics}, 43\penalty0 (6), 2024{\natexlab{c}}.

\bibitem[Yang et~al.(2023)Yang, Zhou, Liu, and Loy]{rerender}
Shuai Yang, Yifan Zhou, Ziwei Liu, and Chen~Change Loy.
\newblock Rerender a video: Zero-shot text-guided video-to-video translation.
\newblock In \emph{SIGGRAPH Asia 2023 Conference Papers}, pages 1--11, 2023.

\bibitem[Yang et~al.(2024{\natexlab{a}})Yang, Hou, Huang, Ma, Wan, Zhang, Chen, and Liao]{direct_a_video}
Shiyuan Yang, Liang Hou, Haibin Huang, Chongyang Ma, Pengfei Wan, Di Zhang, Xiaodong Chen, and Jing Liao.
\newblock Direct-a-video: Customized video generation with user-directed camera movement and object motion.
\newblock In \emph{ACM SIGGRAPH 2024 Conference Papers}, pages 1--12, 2024{\natexlab{a}}.

\bibitem[Yang et~al.(2024{\natexlab{b}})Yang, Teng, Zheng, Ding, Huang, Xu, Yang, Hong, Zhang, Feng, et~al.]{cogvideox}
Zhuoyi Yang, Jiayan Teng, Wendi Zheng, Ming Ding, Shiyu Huang, Jiazheng Xu, Yuanming Yang, Wenyi Hong, Xiaohan Zhang, Guanyu Feng, et~al.
\newblock Cogvideox: Text-to-video diffusion models with an expert transformer.
\newblock \emph{arXiv preprint arXiv:2408.06072}, 2024{\natexlab{b}}.

\bibitem[Yin et~al.(2023)Yin, Wu, Liang, Shi, Li, Ming, and Duan]{dragnuwa}
Shengming Yin, Chenfei Wu, Jian Liang, Jie Shi, Houqiang Li, Gong Ming, and Nan Duan.
\newblock Dragnuwa: Fine-grained control in video generation by integrating text, image, and trajectory.
\newblock \emph{arXiv preprint arXiv:2308.08089}, 2023.

\bibitem[YU et~al.(2025)YU, Hu, Xing, and Shan]{yu2025trajectorycrafter}
Mark YU, Wenbo Hu, Jinbo Xing, and Ying Shan.
\newblock Trajectorycrafter: Redirecting camera trajectory for monocular videos via diffusion models.
\newblock \emph{arXiv preprint arXiv:2503.05638}, 2025.

\bibitem[Zhang et~al.(2024{\natexlab{a}})Zhang, Paiss, Zada, Karnad, Jacobs, Pritch, Mosseri, Shou, Wadhwa, and Ruiz]{recapture}
David~Junhao Zhang, Roni Paiss, Shiran Zada, Nikhil Karnad, David~E Jacobs, Yael Pritch, Inbar Mosseri, Mike~Zheng Shou, Neal Wadhwa, and Nataniel Ruiz.
\newblock Recapture: Generative video camera controls for user-provided videos using masked video fine-tuning.
\newblock \emph{arXiv preprint arXiv:2411.05003}, 2024{\natexlab{a}}.

\bibitem[Zhang et~al.(2024{\natexlab{b}})Zhang, Herrmann, Hur, Jampani, Darrell, Cole, Sun, and Yang]{monst3r}
Junyi Zhang, Charles Herrmann, Junhwa Hur, Varun Jampani, Trevor Darrell, Forrester Cole, Deqing Sun, and Ming-Hsuan Yang.
\newblock Monst3r: A simple approach for estimating geometry in the presence of motion.
\newblock \emph{arXiv preprint arXiv:2410.03825}, 2024{\natexlab{b}}.

\bibitem[Zheng et~al.(2024)Zheng, Li, Jiang, Lu, Wu, and Li]{cami2v}
Guangcong Zheng, Teng Li, Rui Jiang, Yehao Lu, Tao Wu, and Xi Li.
\newblock Cami2v: Camera-controlled image-to-video diffusion model.
\newblock \emph{arXiv preprint arXiv:2410.15957}, 2024.

\bibitem[Zheng et~al.(2025)Zheng, Peng, Zhou, Zhu, Xu, Huang, and Fu]{zheng2025vidcraft3}
Sixiao Zheng, Zimian Peng, Yanpeng Zhou, Yi Zhu, Hang Xu, Xiangru Huang, and Yanwei Fu.
\newblock Vidcraft3: Camera, object, and lighting control for image-to-video generation.
\newblock \emph{arXiv preprint arXiv:2502.07531}, 2025.

\bibitem[Zhou et~al.(2024)Zhou, Yang, Wang, Luo, and Loy]{zhou2024upscale}
Shangchen Zhou, Peiqing Yang, Jianyi Wang, Yihang Luo, and Chen~Change Loy.
\newblock Upscale-a-video: Temporal-consistent diffusion model for real-world video super-resolution.
\newblock In \emph{Proceedings of the IEEE/CVF Conference on Computer Vision and Pattern Recognition}, pages 2535--2545, 2024.

\end{thebibliography}
}

\clearpage
\setcounter{page}{1}
\maketitlesupplementary
\begin{appendices}
\begin{figure*}[t]\centering
\includegraphics[width=0.98\textwidth]{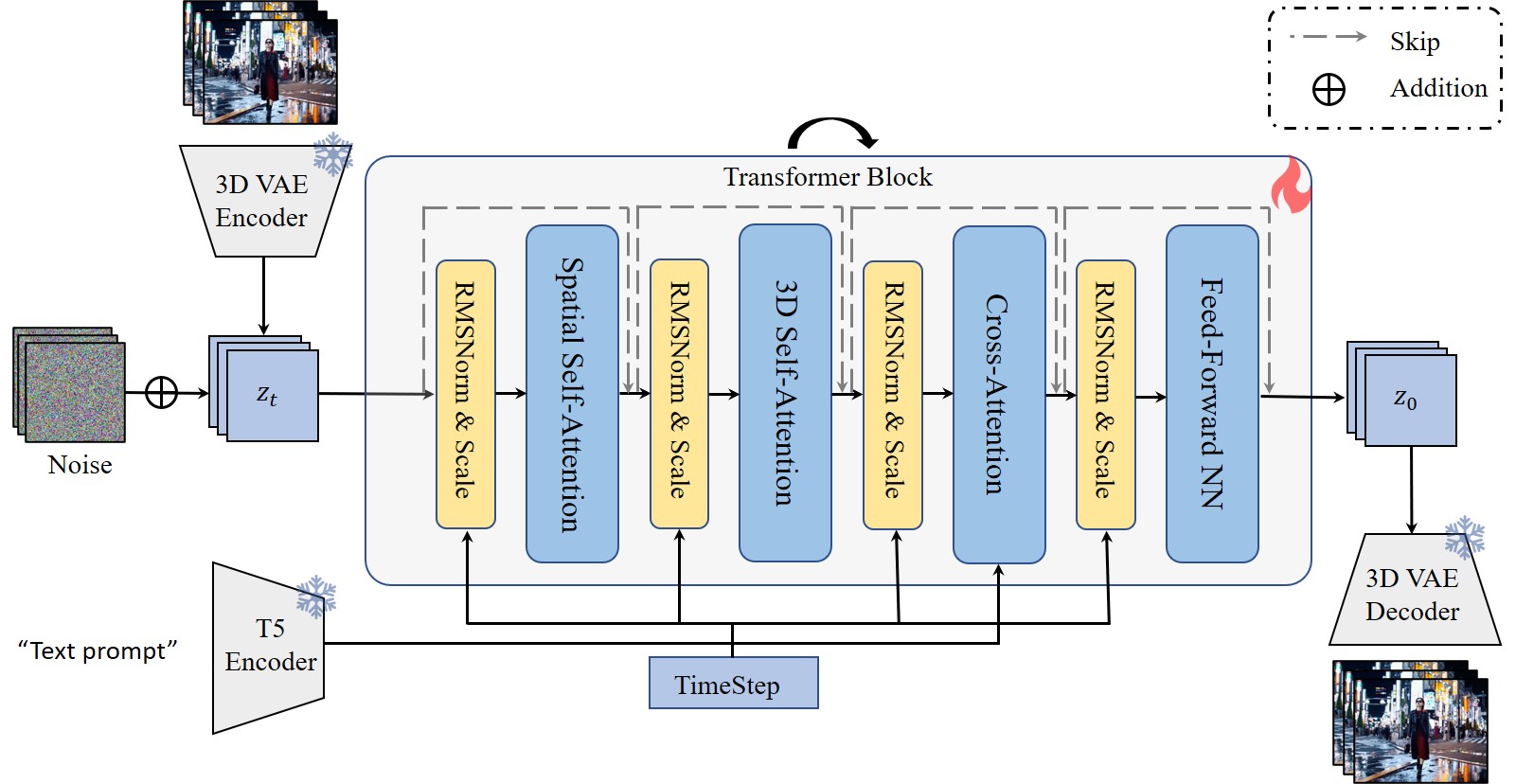}
\caption{\text{Overview of the base text-to-video generation model.}}
    \label{fig_basemodel}
\end{figure*}
\section{Introduction of the Base Text-to-Video Generation Model}
We use a transformer-based latent diffusion model \citep{dit} as the base T2V generation model, as illustrated in Fig. \ref{fig_basemodel}. We employ a 3D-VAE to transform videos from the pixel space to a latent space, upon which we construct a transformer-based video diffusion model. Unlike previous models that rely on UNets or transformers, which typically incorporate an additional 1D temporal attention module for video generation, such spatially-temporally separated designs do not yield optimal results. We replace the 1D temporal attention with 3D self-attention, enabling the model to effectively perceive and process spatiotemporal tokens, thereby achieving a high-quality and coherent video generation model. Specifically, before each attention or feed-forward network (FFN) module, we map the timestep to a scale, thereby applying RMSNorm to the spatiotemporal tokens.
\section{Details of Data Construction}
\label{sec:appendix_data_collection}
In this section, we provide a detailed description of the rendered dataset used to train ReCamMaster.
\vspace{-0.3cm}
\paragraph{3D Environments} We collect 40 different 3D environments assets from \url{https://www.fab.com/}. To minimize the domain gap between rendered data and real-world videos, we primarily select visually realistic 3D scenes, while choosing a few stylized or surreal 3D scenes as a supplement. To ensure data diversity, the selected scenes cover a variety of indoor and outdoor settings, such as city streets, shopping malls, cafes, office rooms, and the countryside.
\vspace{-0.3cm}
\paragraph{Characters} We collected 70 different human 3D models as characters from \url{https://www.fab.com/} and \url{https://www.mixamo.com/#/}, including realistic, anime, and game-style characters.
\vspace{-0.3cm}
\paragraph{Animations} We collected approximately 100 different animations from \url{https://www.fab.com/} and \url{https://www.mixamo.com/#/}, including common actions such as waving, dancing, and cheering. We used these animations to drive the collected characters and created diverse datasets through various combinations.
\vspace{-0.3cm}
\paragraph{Camera Trajectories} 
Due to the wide variety of camera movements, amplitudes, shooting angles, and camera parameters in real-world videos, we need to create as diverse camera trajectories and parameters as possible to cover various situations. To achieve this, we designed some rules to batch-generate random camera starting positions and movement trajectories:

1. Camera Starting Position.

We take the character's position as the center of a hemisphere with a radius of 10m and randomly sample within this range as the camera's starting point, ensuring the closest distance to the character is greater than 0.5m and the pitch angle is within 45 degrees.

2. Camera Trajectories.

\begin{itemize}
    \item \textbf{Pan \& Tilt:} The camera rotation angles are randomly selected within the range, with pan angles ranging from 5 to 60 degrees and tilt angles ranging from 5 to 45 degrees, with directions randomly chosen left/right or up/down.
    
    \item \textbf{Basic Translation:} The camera translates along the positive and negative directions of the xyz axes, with movement distances randomly selected within the range of $[\frac{1}{4}, 1] \times \text{distance2character}$.
    
    \item \textbf{Basic Arc Trajectory:} The camera moves along an arc, with rotation angles randomly selected within the range of 5 to 60 degrees.
    
    \item \textbf{Random Trajectories:} 1-3 points are sampled in space, and the camera moves from the initial position through these points as the movement trajectory, with the total movement distance randomly selected within the range of $[\frac{1}{4}, 1] \times \text{distance2character}$. The polyline is smoothed to make the movement more natural.
    
    \item \textbf{Static Camera:} The camera does not translate or rotate during shooting, maintaining a fixed position.
\end{itemize}

3. Camera Movement Speed.

To further enhance the richness of trajectories and improve our model's generalization ability, 50\% of the training data uses constant-speed camera trajectories, while the other 50\% uses variable-speed trajectories generated by nonlinear functions. Consider a camera trajectory with a total of $f$ frames, starting at location $L_{start}$ and ending at position $L_{end}$. The location at the $i$-th frame is given by:
\begin{equation}
L_i = L_{start} + (L_{end} - L_{start}) \cdot \left( \frac{1 - \exp(-a \cdot i/f)}{1 - \exp(-a)} \right),
\end{equation}
where $a$ is an adjustable parameter to control the trajectory speed. When $a > 0$, the trajectory starts fast and then slows down; when $a < 0$, the trajectory starts slow and then speeds up. The larger the absolute value of $a$, the more drastic the change.

4. Camera Parameters.

We chose two sets of commonly used camera parameters: focal=35mm, aperture=2.8, and focal=24mm, aperture=10.

\begin{figure*}[h]\centering
\includegraphics[width=1\textwidth]{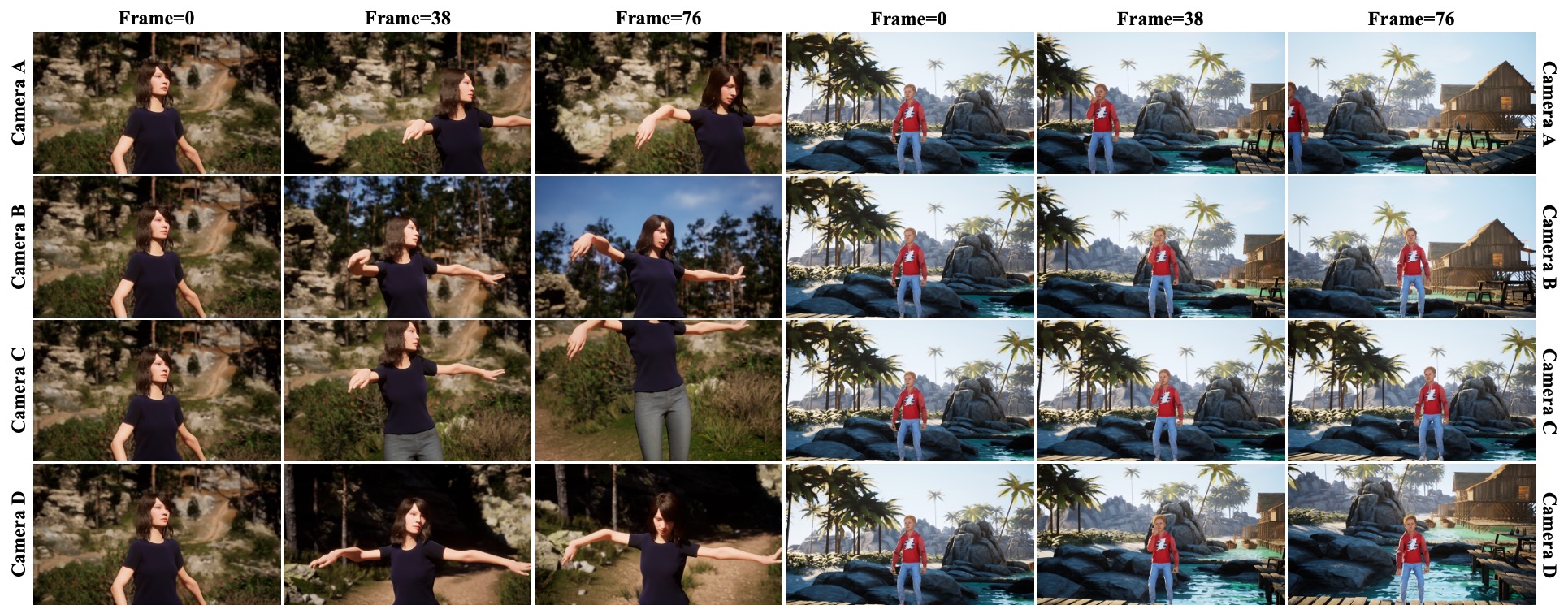}
\caption{Rendered multi-camera synchronized dataset.}
    \vspace{0cm}
    \label{fig_render_data_appendix}

\end{figure*}

\vspace{-0.3cm}
\section{More Results}
\label{sec:appendix_results}
\begin{table*}[!h]
	\begin{center}
		\caption{Ablation study on training data construction.}
            \vspace{-0.00cm}
		\label{tab_ablation_data}
		\setlength\tabcolsep{3.5pt}
		\begin{tabular}{lcccc|cc|ccc}
			\toprule
			\multirow{4}{*}{Dataset} & \multicolumn{4}{c|}{Visual Quality} & \multicolumn{2}{c|}{Camera Accuracy} & \multicolumn{3}{c}{View Synchronization}\\ 
                \cmidrule(r){2-10}
                & \makecell[c]{FID $\downarrow$} & \makecell[c]{FVD $\downarrow$} & \makecell[c]{CLIP-T $\uparrow$} & \makecell[c]{CLIP-F $\uparrow$} & RotErr $\downarrow$& TransErr $\downarrow$ & {Mat. Pix.(K) $\uparrow$} & {FVD-V $\downarrow$} & {CLIP-V $\uparrow$}\\
                \midrule 
                Toy Data & 69.35 & 179.22 & 34.28 & \textbf{98.77} & 1.98 & 5.24 & 862.59 & 89.58 & 89.70 \\
                High-Quality Data & \textbf{57.10} & \textbf{122.74} & \textbf{34.53} & 98.74 & \textbf{1.22} & \textbf{4.85} & \textbf{906.03} & \textbf{90.38} & \textbf{90.36} \\
			\bottomrule
		\end{tabular}
	\end{center}
        \vspace{0cm}
\end{table*}
\subsection{Ablation on Dataset Construction}
\label{sec:appendix_data_ablation}
In our experiments, we find that constructing a diverse training dataset that closely resembles the distribution of real-world videos significantly enhances the model's generalization ability. To demonstrate this, we quantitatively compared the model performance trained on the ``toy data" constructed in the early stages of the experiment and the ``high-quality data" used in this paper. Specifically, the ``toy data" was constructed with 500 scenes in a single 3D environment, and we manually created 20 camera trajectories assigned to 5000 cameras. As a result, this dataset lacked diversity in terms of scenes and camera trajectories, which limited the model's generalization ability on in-the-wild videos and novel camera trajectories. In contrast, the ``high-quality data" used in the paper contains 136K videos shot from 13.6K different dynamic scenes in 40 3D environments with 122K different camera trajectories. We present the results in Tab. \ref{tab_ablation_data}. It is observed that the model shows significant improvements in visual quality, camera accuracy, and synchronization metrics.
\subsection{More Results of ReCamMaster}
\begin{figure}[t]\centering
\includegraphics[width=0.47\textwidth]{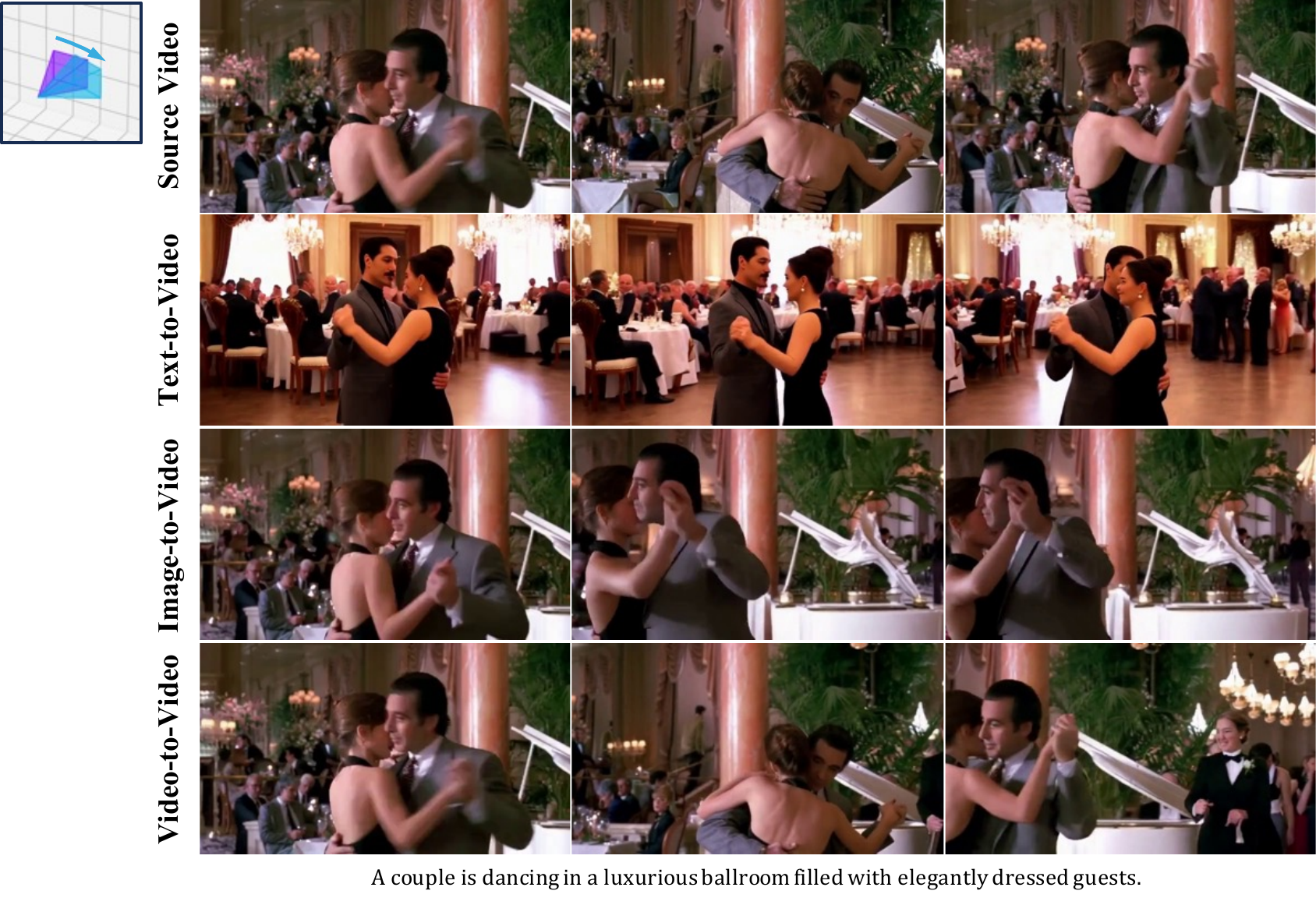}
\caption{Unify camera-controlled tasks with ReCamMaster. ReCamMaster supports T2V, I2V, and V2V camera-controlled generation.}
\vspace{-0.3cm}
    \label{fig_unify_showcase}
\end{figure}

\begin{figure*}[t]\centering
\includegraphics[width=1\textwidth]{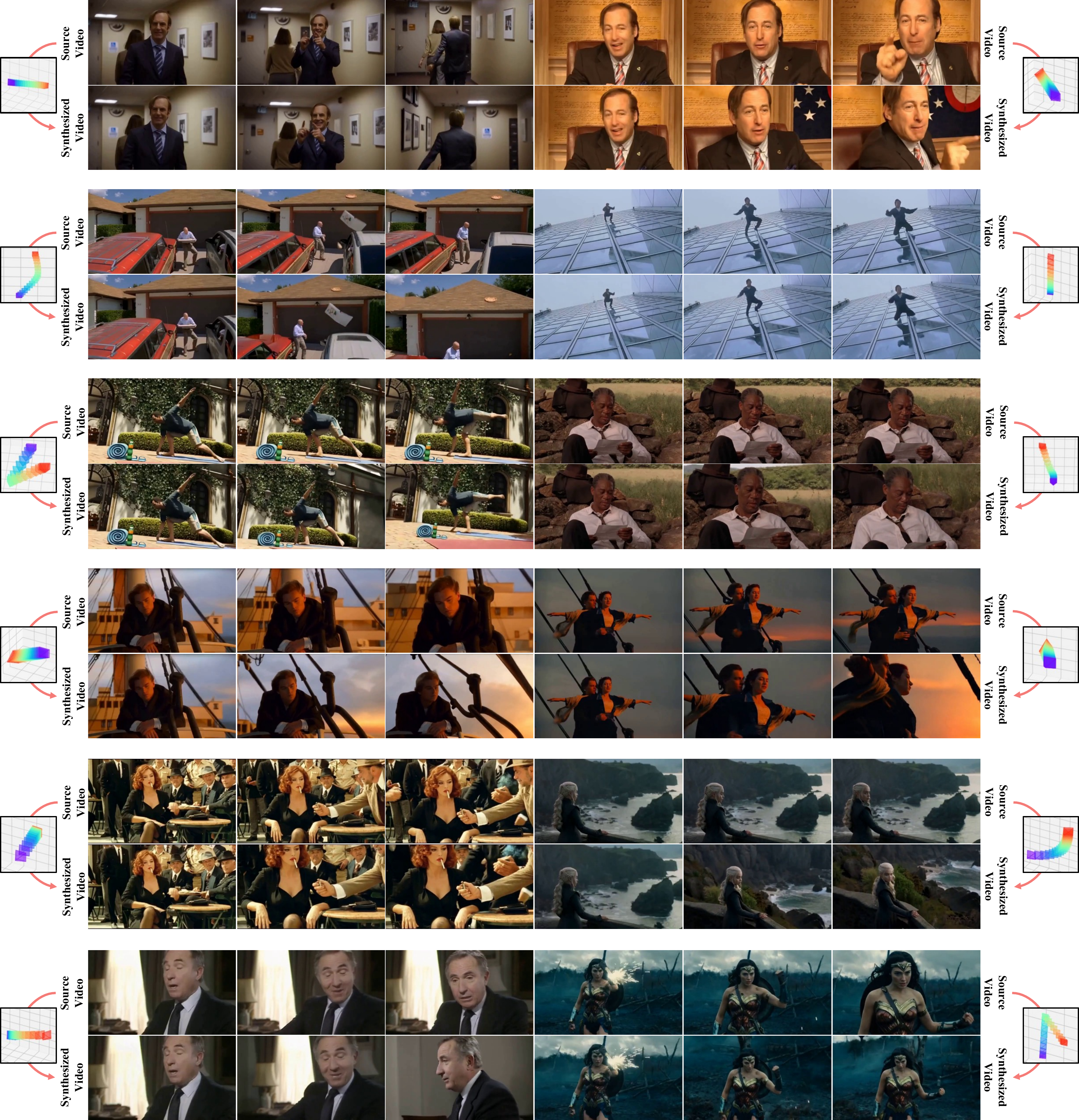}
\caption{More synthesized results of ReCamMaster.}
    \label{fig_appendix_ours1}
\end{figure*}

More synthesized results of ReCamMaster are presented in Fig. \ref{fig_appendix_ours1}. Please visit our \href{https://jianhongbai.github.io/ReCamMaster/}{project page} for more results. We also showcase the ability of ReCamMaster to support T2V, I2V, V2V camera-controlled tasks in Fig. \ref{fig_unify_showcase}. 
\begin{figure}[t]\centering
\includegraphics[width=0.46\textwidth]{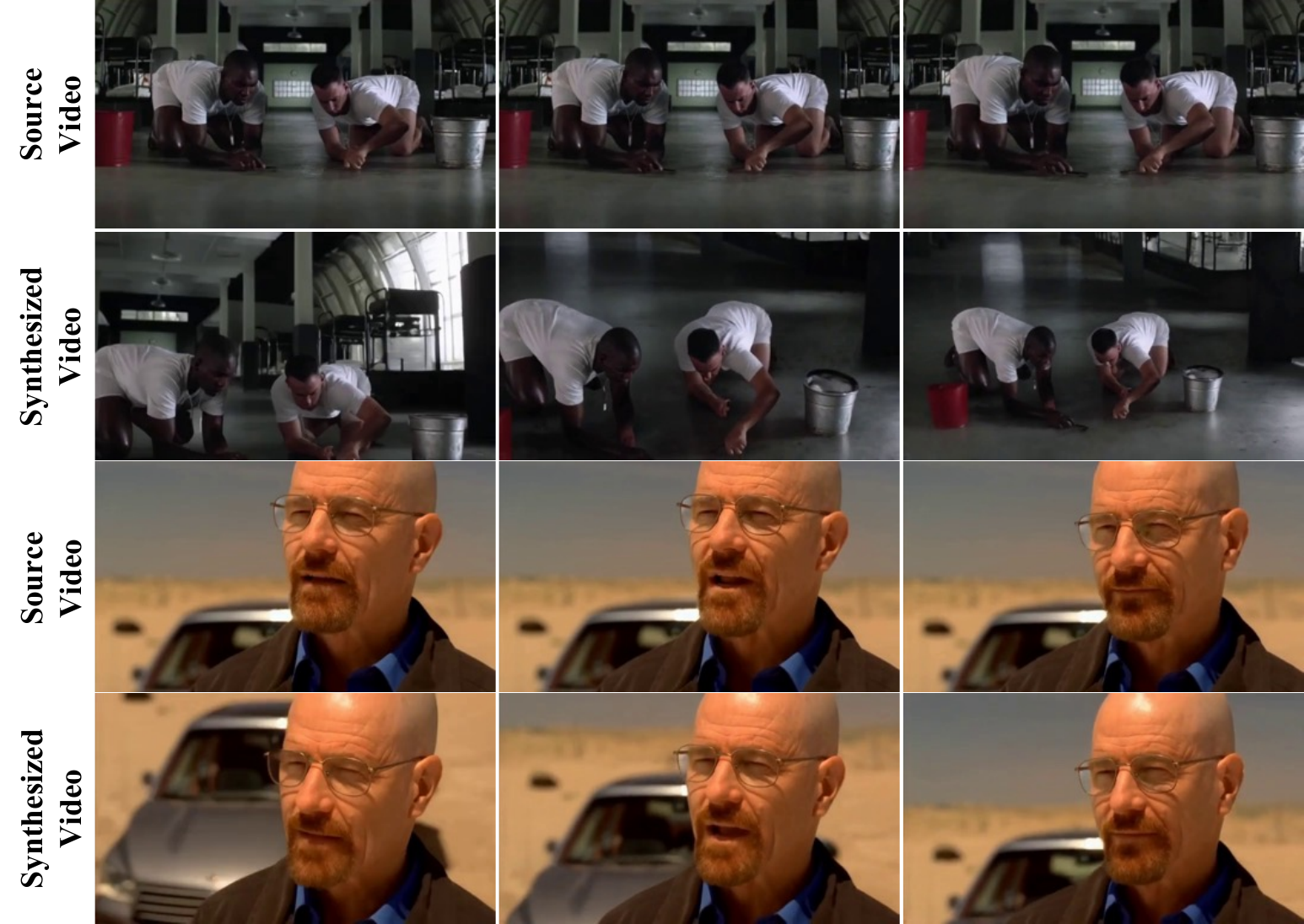}
\vspace{-0.2cm}
\caption{Results on non-overlapping first frames.}
    \label{fig_nonoverlap}
\end{figure}

\begin{figure}[t]\centering
\includegraphics[width=0.46\textwidth]{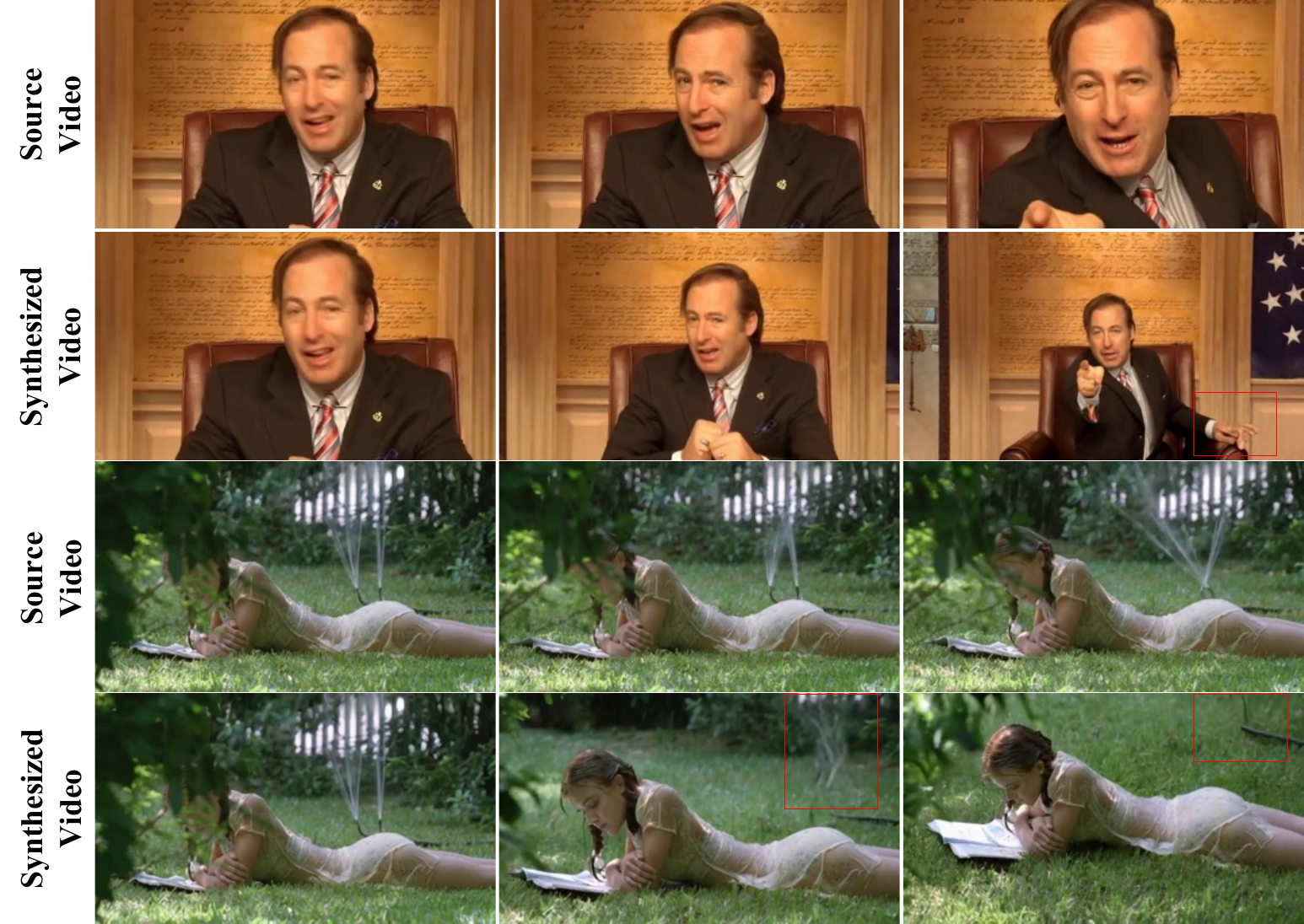}
\vspace{-0.2cm}
\caption{Visualization of failure cases.}
    \label{fig_failure}
    \vspace{-0.4cm}
\end{figure}

\subsection{More Comparison with SOTA Methods}
\begin{figure*}[t]\centering
\includegraphics[width=1\textwidth]{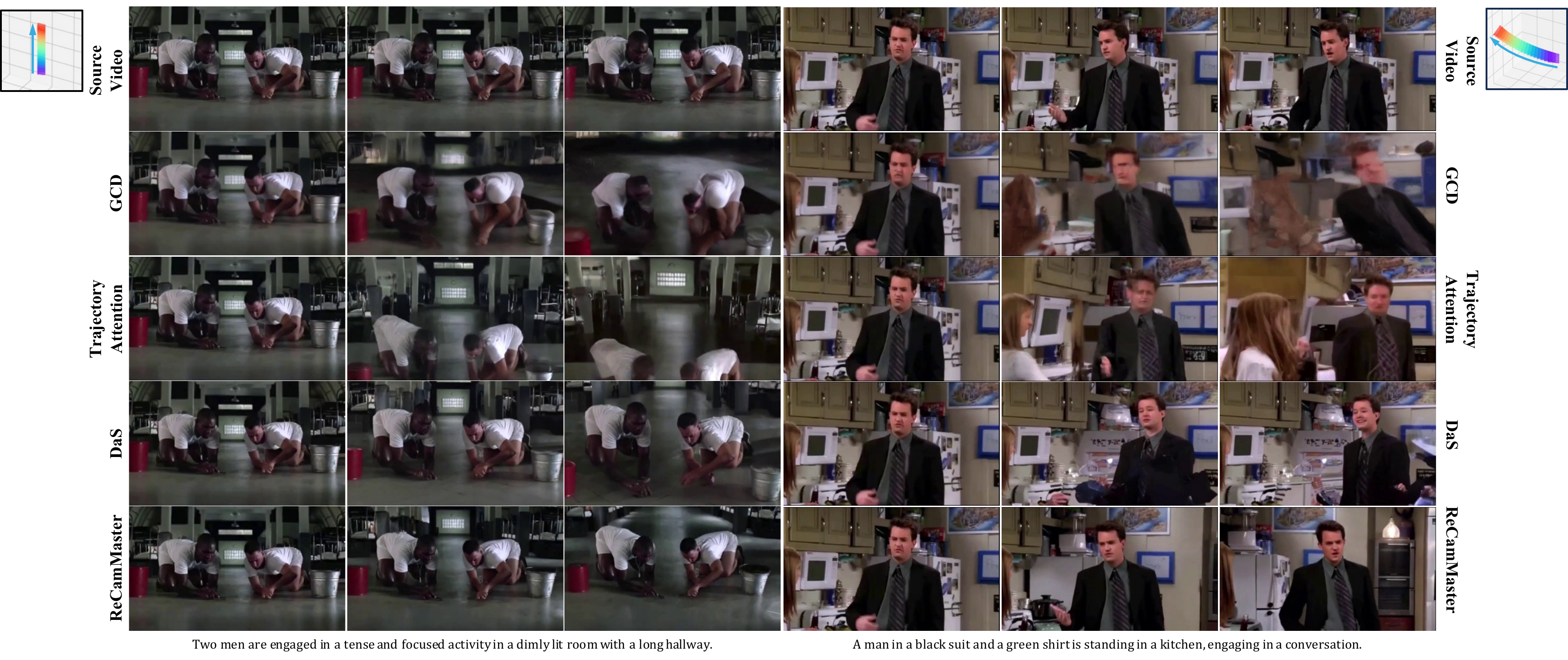}
\caption{More comparison with state-of-the-art methods.}
    \label{fig_appendix_compare1}
\end{figure*}

Please refer to Fig. \ref{fig_appendix_compare1} for qualitative comparison with the state-of-the-art methods.

\subsection{Results on Non-overlapping First Frames}
In the main text, we assume that the first frame of the generated video coincides with the first frame of the input video when testing ReCamMaster. This means the generated video starts from the original video's first frame. To evaluate the method's generalizability, we also rendered 27K `non-overlapping first-frame' videos as training data and tested whether the generated first frame could differ from the original. Fig. \ref{fig_nonoverlap} presents the qualitative results, with the leftmost column showing the first frame. It is evident that the model generalizes well to non-overlapping first frames, enabling re-filming from a completely new perspective, thus demonstrating the method's generalizability.

\subsection{Failure Cases Visualization}
We present the failure cases in Fig. \ref{fig_failure}. Since our model is built upon a text-to-video base model, we also inherit some of the base model's shortcomings. For instance, the generated hand movements of characters may exhibit inferior quality, as shown in the first and second rows. Moreover, generating very small objects sometimes results in failures, as shown in the third and fourth rows.
\end{appendices}

\end{document}